\documentclass[twoside,11pt]{article}
\usepackage[abbrvbib, preprint]{jmlr2e}

\usepackage[utf8]{inputenc}
\usepackage[a4paper,left=2.5cm,right=2.5cm,top=3cm,bottom=4cm]{geometry}
\usepackage{verbatim}
\usepackage{amsmath}
\usepackage{enumitem}
\usepackage{dsfont}
\usepackage{mathtools}
\usepackage[dvipsnames]{xcolor}
\usepackage{tikz-cd} 
\usepackage{mdframed}
\usepackage{algpseudocode}
\usepackage{overpic}

\newcommand{\dd}{\mathrm d}
\newcommand{\gl}{\operatorname{GL}}

\newtheorem{theoremMain}{Theorem}

\newtheorem{hypothesis}{Hypothesis}

\newcommand{\revision}[1]{{#1}}

\numberwithin{equation}{section}
\numberwithin{theorem}{section}
\usepackage{lastpage}

\ShortHeadings{Characterizing Dynamic Stability of SGD}{Chemnitz and Engel}
\firstpageno{1}

\title{Characterizing Dynamical Stability of Stochastic Gradient Descent in Overparameterized Learning}
\author{\name Dennis Chemnitz \email dennis.chemnitz2@fu-berlin.de\\ \addr Fachbereich Mathematik und Informatik\\ Freie Universität Berlin \\ 14195 Berlin, Germany \AND \name Maximilian Engel \email m.r.engel@uva.nl \\ \addr KdV Institute for Mathematics\\
University of Amsterdam \\ 1090 GE Amsterdam, Netherlands\\ \textnormal{and} \\  Fachbereich Mathematik und Informatik\\ Freie Universität Berlin \\ 14195 Berlin, Germany}
\date{\today}

\begin{document}
\maketitle
\begin{abstract}
    For overparameterized optimization tasks, such as those found in modern machine learning, global minima are generally not unique. In order to understand generalization in these settings, it is vital to study to which minimum an optimization algorithm converges. The possibility of having minima that are unstable under the dynamics imposed by the optimization algorithm limits the potential minima that the algorithm can find. In this paper, we characterize the global minima that are dynamically stable/unstable for both deterministic and stochastic gradient descent (SGD). 
    In particular, we introduce a characteristic Lyapunov exponent that depends on the local dynamics around a global minimum and rigorously prove that the sign of this Lyapunov exponent determines whether SGD can accumulate at the respective global minimum.
\end{abstract}
\section{Introduction}
Since the success of ``AlexNet'' \citep{alexnet}, the overparameterization paradigm has become ubiquitous in modern machine learning. Although it is well known that artificial neural networks with sufficiently many parameters can approximate arbitrary goal functions (see, e.g.,~\citealp{Approx}), the training process, especially for overparameterized networks, is not well understood and leaves many open questions. For one, the fact that the loss landscapes are usually non-convex makes it difficult to rigorously guarantee that the optimization algorithms converge to global minima. 

Another key open problem in understanding the training of overparameterized networks is that of generalization (see, e.g.,~\citealp{generalization}). Classical wisdom suggests that overparameterized networks should suffer from overfitting problems due to their high expressivity. However, in practice, this is not the case. 
In other words, sufficiently large (i.e.~deep, wide, or both) artificial neural networks are capable of fitting arbitrary data, due to their large number of parameters (typically weights and biases).
Thus, finding a set of parameters for which the network interpolates the training data, in itself, does not indicate that the network makes reasonable predictions for inputs outside the training data set. 
However, the parameters found by common optimization algorithms tend to perform well on unseen data. This is essential for the success of modern machine learning.

One possible explanation for this phenomenon is that of dynamical stability (see, e.g.~\citealp{WuMaE}). Since the learning rate is in practice not infinitesimally small, some global minima can become dynamically unstable, in the sense that the optimization algorithm will not converge to these solutions, even if it is initialized arbitrarily close to them. This excludes unstable optimal solutions from being found by the optimization algorithm, thereby reducing the effective hypothesis class to stable global minima. 

For gradient descent, a second-order Taylor expansion shows that a global minimum is only stable if the largest eigenvalue of the Hessian of the empirical loss is less than $2/\eta$, where $\eta$ is the learning rate. Thus, gradient descent can only converge to minima that are sufficiently flat, which has been associated with good generalization (see e.g.~\citealp{flatmininma}). 
Moreover, numerical experiments by \citet{Cohen22} have demonstrated that (deterministic) gradient descent operates at the \emph{edge of stability}, which means that the largest eigenvalue of the Hessian oscillates around the critical value $2/\eta$ for the later stages of training.

Transferring these ideas to the stochastic case requires an appropriate notion of dynamical stability under stochastic gradient descent (cf.~\citealp{WuMaE} and \citealp{maYing}). The goal of this work is to establish a coherent mathematical framework in which the dynamical stability of deterministic and stochastic optimization algorithms can be investigated and the convergence properties of global minima can be characterized.

\subsection{Contributions}\label{sec:contributions}
We investigate which global minima of the empirical loss function can be obtained as limits of the two most fundamental optimization algorithms, namely gradient descent (GD) and stochastic gradient descent (SGD). Specifically, we consider the limits of GD and SGD as random variables, denoted by $X_{\lim}^{\operatorname{GD}}$ and $X_{\lim}^{\operatorname{SGD}}$, respectively. These random variables depend on the randomness during initialization in the case of GD and on the randomness during initialization and training in the case of SGD. For a given global minimum $x^*$ of the empirical loss $\mathcal L$, we introduce the quantity 
$$\mu(x^*):= \log\|\mathds 1 - \eta \operatorname{Hess} \mathcal L(x^*)\|.$$
As is readily seen in Section \ref{sec:linstab}, the above condition $\|\operatorname{Hess} \mathcal L (x^*)\|< \frac{2}{\eta}$ is equivalent to $\mu(x^*)< 0$. We choose this reformulation because it has a direct analogue for SGD.

Our main contributions are the following.
\begin{itemize}
    \item We introduce a set of mild conditions (cf.~\ref{assu:manifold}, \ref{assu:init}, \ref{assu:nonsing} below) under which we can rigorously prove that the sign of $\mu$ characterizes the support of $X_{\lim}^{\operatorname{GD}}$ (cf.~Theorem \ref{theo:mainGD}).
    \item For global minima $x^*$, we introduce a new quantity, denoted by $\lambda(x^*)$ (cf.~\eqref{eq:lambdaDef} below), which can be seen as a characteristic Lyapunov exponent \citep{Osdeledec68} and corresponds to a new notion of dynamic stability/instability for SGD. It should be noted that this notion differs\footnote{In particular our notion of linear stability is strictly weaker than the one introduced in \cite[Defintion 1]{WuMaE} (cf.~Appendix \ref{app:WuMaW}).} from the ones introduced in \cite{WuMaE} and \cite{maYing} (cf.~Appendix \ref{app:WuMaW}).
    \item Under an additional mild assumption on the global minimum in question, we rigorously prove that the sign of $\lambda$ characterizes the support of $X_{\lim}^{\operatorname{SGD}}$ (cf.~Theorem \ref{theo:mainSGD}).
\end{itemize}

This clarifies the concept of dynamical stability for stochastic gradient descent and establishes a connection to a substantial body of mathematical and physical literature on Lyapunov exponents (see, e.g.~\citealp{Osdeledec68,RMP, Bible}). Moreover, it provides a foundation for new theoretic and numerical investigations of the learning process under stochastic gradient descent. Possible questions of interest include how the value of $\lambda$ evolves during training with stochastic gradient descent and what implications a small, respectively large, value of $\lambda$ has on generalization.

From a mathematical perspective, our analysis of SGD amounts to a study of the asymptotic behavior of discrete-time random dynamical systems (see, e.g.~\citealp{Bible} and also Section \ref{sec:RDS}) possessing a manifold of equilibria. In the unstable case, we use a method based on moment Lyapunov exponents (see, e.g.~\citealp{ArnoldKliemannOeljeklaus86}). A more detailed discussion of this approach is given in Section \ref{sec:proofOverview} and throughout the proofs.

For simplicity, we restrict our study to scalar regression problems and, in the case of SGD, to mini-batch size 1, but we point out that, with minor modifications, our main result applies to a broader class of tasks and algorithms (cf.~Section \ref{sec:open} below). 
\subsection{Related Literature}
 Closest to our work is a series of two papers by \cite{WuMaE} and \cite{maYing}, which study the dynamical stability of stochastic gradient descent. We explain the relation to this work in detail in Appendix \ref{app:WuMaW}. 
 
 \cite{gurbuzbalaban21} consider Lyapunov exponents of SGD in a setting where SGD does not converge to a global minimum, but the transition densities approach a heavy-tailed stationary distribution (see also e.g.~\citealp{hodgkinson21}). The convergence rates for SGD with decaying learning rate have been studied in \cite{FGJ20}.
 
The edge of stability phenomenon for gradient descent mentioned in the introduction has been numerically observed by \cite{WuMaE}, as well as by \cite{Cohen22} and theoretically studied by \cite{arora22}. Moreover, the difference between the global minima found by GD and SGD (see \citealp{keskar2017b, WuMaE}) and between SGD and Adam (see \citealp{Keskar17}) has been numerically investigated.

 The dynamics of stochastic gradient descent have also been studied using approximations by stochastic differential equations, called stochastic modified equations (see, e.g.~\cite{Li17,Li19}). We note that such an approach is not compatible with our analysis as it assumes an infinitesimally small learning rate. Moreover, the modified stochastic equations have been extended to a stochastic flow in \cite{Gess24}, building a framework analogous to the one we introduce in Section \ref{sec:RDS}.
 
The most critical part of our proof is based on techniques originally developed to study synchronization in stochastic differential equations by \cite{BaxendaleStroock} (see also \citealp{Baxendale1991}), which has received recent attention motivated by fluid dynamics \citep{CotiZelatiHairer,BedrossianBlumenthalPunshon2022,bedrossian2025,BlumenthalCotiGvalani2023}.
\subsection{Organization of the Paper}
We start by introducing the fundamental setting of the optimization task (Section \ref{sec:nn}), as well as the learning algorithms that we consider (\ref{subs:sgd}). In Section \ref{sec:linstab}, we heuristically derive the notions of linear stability for GD and SGD. These derivations are made rigorous in our main results, which we present in Section \ref{sec:mainRes}.
In Section \ref{sec:ntk}, the main quantities of interest, $\mu$ and $\lambda$, are related to the neural tangent kernel (cf.~\citealp{NTK}).
Finally, we discuss possible generalizations of our main results\revision{, such as extensions to other optimizers,} in Section \ref{sec:open}.
Section \ref{sec:proofs} provides the proofs of the main theorems. An overview of the structure of the proofs is given in Section \ref{sec:proofOverview}.

\section{Setting and Main Result}
\subsection{Network Model}\label{sec:nn}
In the following, we consider a scalar regression problem. Let $\hat f: \mathbb R^d \to \mathbb R$ be a ground truth function, which is supposed to be reconstructed from $N$ given data pairs $(y_i, z_i = \hat f(y_i))_{i \in [N]}$, where $[N]= \{1, \dots, N\}$. To do so, we consider a parameterized network model given by a smooth function\footnote{For the purpose of being general, we will not specify how the network function $\mathfrak F$ looks like. Typical network architectures used in practice include fully connected networks, convolutional neural networks (image classification), and transformers (large language models).} $\mathfrak F: \mathbb R^D \times \mathbb R^d \to \mathbb R$ and try, using the given data, to find an $x \in \mathbb R^D$ such that $\mathfrak F (x, \cdot) \approx \hat f$. In the absence of other information, it is reasonable to prescribe 
\begin{equation}\label{eq:xinargmin}
    x \in \operatorname{argmin} \frac{1}{N} \sum_{i \in [N]}\ell(\mathfrak F(x, y_i), z_i),
\end{equation}
for some loss function $\ell: \mathbb R\times \mathbb R \to \mathbb R_{\geq 0}$, with $\ell(\bar z,z) = 0$ if and only if $\bar z = z$, $\partial_1 \ell(z,z) = 0$ and $\partial_1^2 \ell( z, z) = 1$. Although other choices for $\ell$ are possible, it is recommended to think of the square loss function $\ell(\bar z, z) := \frac{1}{2}(\bar z - z)^2$. For each $i \in [N]$, we define the \emph{individual loss function} $\mathcal L_i : \mathbb R^D \to [0, \infty)$ by
\begin{equation} \label{eq:indiv_loss}
   \mathcal  L_i(x) := \ell (\mathfrak F(x,y_i), z_i)
\end{equation}
and the \emph{empirical loss function} $\mathcal L : \mathbb R^D \to \mathbb R$ as the average of the individual loss functions, that is, 
\begin{equation} \label{eq:emp_loss}
\mathcal L(x) := \frac{1}{N} \sum_{i = 1}^N \mathcal L_i(x).
\end{equation}
Equation \eqref{eq:xinargmin} can thus be rewritten as
\begin{equation}\label{eq:xinargmin2}
    x \in \operatorname{argmin} \mathcal L(x).
\end{equation}
If the number of given training examples $N$ exceeds the number of learnable parameters $D$, the problem is overdetermined and the global minimum of $\mathcal L$ is usually unique, so that $x$ is fully determined by \eqref{eq:xinargmin2}. However, in general, it will not be possible to find an $x \in \mathbb R^D$ such that $\mathfrak F(x, \cdot)$ interpolates the given data, i.e.~the set 
$$\mathcal M:= \left\{x\in \mathbb R^D : \mathcal L(x) = 0\right\} = \left\{x \in \mathbb R^D: \mathfrak F(x, y_i) = z_i, ~\forall i \in [N]\right\}$$
will be empty.

On the other hand, if the number of learnable parameters exceeds the number of training examples, i.e.~$D>N$, the optimization problem is overparameterized. As described in the introduction, this will be the case of interest here.
Given a sufficiently expressive network model $\mathfrak F$ (meaning that $\mathfrak F(x,\cdot)$ can express a sufficiently rich family of functions), the set of interpolation solutions $\mathcal M$ will usually be infinite. In fact, if the set of gradients
$$\left\{\nabla_x \mathfrak F(x,y_i) : i \in [N]\right\} \subset R^D$$
is linearly independent for every $x \in \mathcal M$, the set $\mathcal M$ is an embedded smooth $(D-N)$-dimensional submanifold of $\mathbb R^D$ with normal space
\begin{equation} \label{eq:normal_space}
\mathcal N(x) = \operatorname{span}\{\nabla_x \mathfrak F(x,y_i): i \in [N]\}
\end{equation}
and tangent space
\begin{equation} \label{eq:tangent_space}
\mathcal T(x) = \mathcal N(x)^\perp = \{v \in \mathbb R^D : w^t v = 0, ~ \forall \,w \in  \mathcal N(x)\}.
\end{equation}
As a consequence of Sard's theorem (see e.g.~\citealp{Milnor65}), this condition will be satisfied for generic training 
data\footnote{Put more precisely, for every smooth network model $\mathfrak F$ and every choice of $(y_1, \dots, y_N)$, there exists a set $\mathcal Z \subseteq \mathbb R^N$ of full Lebesgue measure, such that, if $(z_1, \dots, z_N) \in \mathcal Z$, then $\mathcal M$ is an embedded smooth $(D-N)$-dimensional manifold (cf.~\citealp{YCooper}).}  and in the following we will assume that $\mathcal M$ is a smooth manifold as described (cf.~Hypothesis \ref{assu:manifold} below). 

While $x \in \mathcal M$ ensures $\mathfrak F(x, y_i) = \hat f(y_i)$ for all $i \in [N]$, it does not guarantee $\mathfrak F(x, y) \approx \hat f(y)$ for any $y$ outside the training data set. This is the generalization gap mentioned in the introduction. If $\mathcal M$ is infinite, \eqref{eq:xinargmin2} does not determine $x$ uniquely and the question of generalization is highly dependent on which $x \in \mathcal M$ is chosen. 

\subsection{Learning Algorithms}\label{subs:sgd}
In practice, a wide range of optimization algorithms (cf.~e.g.~\citealp{Schmidt21}) are used to find a global minimum $x^* \in \mathcal M$.
In this work, we will restrict our study to the two most basic optimization algorithms: gradient descent (GD) and stochastic gradient descent (SGD). Both algorithms are initialized at a random point $X_0^{\operatorname{GD/SGD}} \in \mathbb R^D$, which is distributed according to some probability measure $\nu: \mathfrak B(\mathbb R^D) \to [0,1]$, called the \emph{initial distribution}.\footnote{Here and in the following, $\mathfrak B(\mathbb R^D)$ denotes the family of Borel sets.}
We will assume that $\nu$ is equivalent to Lebesgue measure\footnote{This is for example consistent with $\nu$ being a normal distribution, which is the most common choice in practice.} (cf.~Hypothesis \ref{assu:init} below).

Gradient descent tries to successively reduce the empirical loss $\mathcal L$ by taking small steps in the opposite direction of its gradient $\nabla \mathcal L$. Formally, the update rule is given by
\begin{equation}\label{eq:GD}
    X^{\operatorname{GD}}_{n+1} := X^{\operatorname{GD}}_n - \eta \nabla \mathcal L (X^{\operatorname{GD}}_n),
\end{equation}
where $\eta>0$ is a small real number called \emph{learning rate}. Ideally, the hope is that the gradient descent converges to some point ${X^{\operatorname{GD}}_{\lim} \in \mathcal M}$ as $n$ tends to infinity. Although gradient descent is remarkably reliable in finding global minima in practice, convergence to a global minimum (or even convergence in the first place) cannot be guaranteed in our general setting, since $\mathcal L$ is generally 
non-convex.
Therefore we define $X^{\operatorname{GD}}_{\lim}$ as an $\mathcal M \cup \{\emptyset\}$-valued random variable by 
$$X^{\operatorname{GD}}_{\lim} := \begin{cases}\lim_{n \to \infty} X^{\operatorname{GD}}_n &, \text{ if $(X^{\operatorname{GD}}_n)$ converges to some point in $\mathcal M$,}\\
\emptyset &, \ \text{otherwise,}\end{cases}$$
accounting for a possible failure of convergence. Note that the randomness in $X^{\operatorname{GD}}_{\lim}$ stems purely from the random initialization $X^{\operatorname{GD}}_0$. Once $X^{\operatorname{GD}}_0$ is drawn, all subsequent steps, and thus $X^{\operatorname{GD}}_{\lim}$ are deterministic.

 In contrast, stochastic gradient descent is a stochastic optimization algorithm. After initializing $X^{\operatorname{SGD}}_0$ randomly according to the measure $\nu$, SGD updates are performed according to the gradients of the individual loss functions $\mathcal L_i$ \eqref{eq:indiv_loss}, corresponding to randomly chosen training data. Concretely, we set
\begin{equation}\label{eq:SGD}
    X^{\operatorname{SGD}}_{n+1} :=  X^{\operatorname{SGD}}_n - \eta \nabla \mathcal L_{\xi_{n+1}} (X^{\operatorname{SGD}}_n),
\end{equation}
where $(\xi_n)_{n \in \mathbb N}$ is a sequence of independent random variables, which are uniformly distributed over $[N]$. 
Note that $\nabla \mathcal L_{\xi_{n}}$ can be interpreted as an unbiased estimator for the gradient $\nabla \mathcal L$ of the empirical loss function $\mathcal L$ \eqref{eq:emp_loss} since 
$$\mathbb E [\nabla \mathcal L_{\xi_{n}}(x)] = \nabla \mathcal L(x),$$
for each $x \in \mathbb R^D$ and each $n \in \mathbb N$. Unfortunately, the problems with guaranteeing convergence for SGD persist such that, again, we define $X^{\operatorname{SGD}}_{\lim}$  as a $\mathcal M \cup \{\emptyset\}$-valued random variable by 
$$X^{\operatorname{SGD}}_{\lim} := \begin{cases}\lim_{n \to \infty} X^{\operatorname{SGD}}_n &, \text{ if $(X^{\operatorname{SGD}}_n)$ converges to some point in $\mathcal M$,}\\
\emptyset &, \text{ otherwise.}\end{cases}$$

In the following, we are interested in studying which points GD and SGD can find. Put more precisely, we aim to characterize the support of the random variable $X^{\operatorname{GD/SGD}}_{\lim}$, which is defined as
$$\operatorname{supp} \left(X^{\operatorname{GD/SGD}}_{\lim}\right) := \left\{x \in \mathcal M :\revision{\mathbb P\left(X^{\operatorname{GD/SGD}}_{\lim} \in U\right) >0, ~\forall \,U \subseteq\mathcal M \text{ open nbhd. of } x}\right\}.$$

\subsection{Linear Stability}\label{sec:linstab}
If (stochastic) gradient descent is randomly initialized at some $x \in \mathcal M$ or reaches $\mathcal M$ after some finite number of iterations, it remains stationary from there on, i.e.
$$X^{\operatorname{GD/SGD}}_n \in \mathcal M \Rightarrow X^{\operatorname{GD/SGD}}_n = X^{\operatorname{GD/SGD}}_m = X^{\operatorname{GD/SGD}}_{\lim}, ~\forall\, m > n.$$
This can be seen easily by computing the individual and empirical loss functions as
\begin{align*}
    \nabla \mathcal L_i(x) &= \partial_1\ell(\mathfrak F(x, y_i), z_i) \nabla_x \mathfrak F(x, y_i),\\
    \nabla \mathcal L(x) &= \frac{1}{N} \sum_{i = 1}^N\partial_1\ell(\mathfrak F(x, y_i),z_i) \nabla_x \mathfrak F(x, y_i)
\end{align*}
and noting that the right-hand sides vanish if $x \in \mathcal M$.
Since the initial distribution $\nu$ is assumed to have full support, it is possible that $X^{\operatorname{GD/SGD}}_{\lim}$ takes any value in $\mathcal M$. However, under mild assumptions (cf.~\ref{assu:init} and \ref{assu:nonsing} below), the optimization algorithm can only reach a global minimum in a finite number of steps with probability zero. In other words, (S)GD will usually converge to some global minimum, say $x^* \in \mathcal M$, without reaching it in a finite number of steps. This is only possible if $x^*$ is \emph{dynamically stable}, which means that, once the optimization algorithm gets sufficiently close to $x^*$, it will stay close. 

In order to study the dynamical stability at some $x^* \in \mathcal M$, we \emph{linearize} the optimization step \eqref{eq:GD} around $x^*$.
We introduce the function $\mathcal G_\eta: \mathbb R^D \to \mathbb R^D$ by
\begin{equation}\label{eq:Geta}
    \mathcal G_\eta (x) :=  x - \eta \nabla \mathcal L (x),
\end{equation}
such that $ X^{\operatorname{GD}}_{n+1} =\mathcal G (X^{\operatorname{GD}}_n)$. If $x$ is close to $x^*$, by differentiability, $\mathcal G_\eta(x)$ is well approximated by
$$\mathcal G_\eta(x) = x^* + \mathcal G'_\eta(x^*) (x-x^*) + o(\|x-x^*\|).$$
Here, $\mathcal G'_\eta(x^*) \in \mathbb R^{D\times D}$ denotes the Jacobian of $\mathcal G_\eta$ at $x^*$. Consequently, for several iterations, we have
$$\mathcal G_\eta^n(x) = x^* + \mathcal G'_\eta(x^*)^n (x-x^*) + o(\|x-x^*\|).$$
The Jacobian $\mathcal G'_\eta(x^*)$ can be computed as 
\begin{equation}\label{eq:JacG}
    \mathcal G'_\eta(x^*) = \mathds 1_D - \eta \operatorname{Hess} \mathcal L (x^*) =  \mathds 1_D - \frac{\eta}{N} \sum_{i = 1}^N \nabla_x\mathfrak F(x^*,y_i)\nabla_x\mathfrak F(x^*,y_i)^t.
\end{equation}
Note that, by definition of the normal space $\mathcal N$ \eqref{eq:normal_space} and the tangent space $\mathcal T$ \eqref{eq:tangent_space}, the matrix $\mathcal G'_\eta(x^*)$ satisfies $\mathcal G'_\eta(x^*) v = v \in \mathcal T(x^*)$ for all $v \in \mathcal T(x^*)$ and $\mathcal G'_\eta(x^*) w \in \mathcal N(x^*)$ for all $w \in \mathcal N(x^*)$. 
Thus, $\mathcal G'_\eta(x^*)$ respects the splitting $\mathbb R^D = \mathcal T(x^*) \oplus \mathcal N(x^*)$ and the restriction $\mathcal G'_\eta(x^*)|_{\mathcal T(x^*)}: \mathcal T(x^*) \to \mathcal T(x^*)$ equals the identity.
We say that $x^*$ is \emph{linearly stable} under GD with learning rate $\eta$ if
$$\mu(x^*) := \lim_{n \to \infty}\frac{1}{n}\log(\|\mathcal G'_\eta(x^*)^n|_{\mathcal N(x^*)}\|) = \log(\rho_{\operatorname{Spec}}(\mathcal G'_\eta(x^*)|_{\mathcal N(x^*)})) < 0.$$
Here $\|\cdot\|$ denotes the operator norm and $\rho_{\operatorname{Spec}}$ the spectral radius of a matrix which is given by
$$\rho_{\operatorname{Spec}}(A) := \lim_{n \to \infty} \|A^n\|^{1/n} = \max \{|\lambda| : \lambda \text{ is an eigenvalue of }A\}.$$
Conversely, if $\mu(x^*)> 0$, we say that $x^*$ is \emph{linearly unstable} under GD. By \eqref{eq:JacG}, we have
$$\operatorname{Spec}(\mathcal G'_\eta(x^*)|_{\mathcal N(x^*)}) = 1- \eta \operatorname{Spec}(\operatorname{Hess} \mathcal L (x^*)|_{\mathcal N(x^*)}).$$
While the Hessian $\operatorname{Hess} \mathcal L (x^*)$ is only positive semi-definite, its restriction to $\mathcal N(x^*)$ is even positive definite and thus
$$\operatorname{Spec}(\operatorname{Hess} \mathcal L (x^*)|_{\mathcal N(x^*)}) \subset \mathbb R_{>0}.$$
In particular, the Hessian $\operatorname{Hess} \mathcal L (x^*)|_{\mathcal N(x^*)}$ is a symmetric positive definite matrix.
Therefore, the condition for linear stability $\mu(x^*) < 0$ can be equivalently expressed as
\begin{align*}
    \mu(x^*) < 0 &\Leftrightarrow \operatorname{Spec}(\mathcal G'_\eta(x^*)|_{\mathcal N(x^*)}) \subset (-1,1) \Leftrightarrow \operatorname{Spec}(\operatorname{Hess} \mathcal L (x^*)|_{\mathcal N(x^*)}) \subset \left(0, 2/\eta\right)\\
    &\Leftrightarrow \|\operatorname{Hess} \mathcal L (x^*)|_{\mathcal N(x^*)})\|< \frac{2}{\eta} \Leftrightarrow \|\operatorname{Hess} \mathcal L (x^*)\|< \frac{2}{\eta},
\end{align*}
where the last equivalence holds if $\eta < 2$.  Although this formulation is more common in the literature (cf., e.g.~\citealp{WuMaE,Cohen22,arora22}), we will stick to the expression $\mu(x^*) < 0$ here, as it can be more easily extended to stochastic gradient descent.

Our stability analysis for stochastic gradient descent is analogous to that for gradient descent. For notational convenience, we introduce the functions $\mathcal G_{\eta, i}: \mathbb R^D \to \mathbb R^D$ for each $i \in [N]$ by 
\begin{equation}\label{eq:Getai}
    \mathcal G_{\eta, i}(x) := x-\eta \nabla \mathcal L_i(x).
\end{equation}
Their Jacobians at some global minimum $x^* \in \mathcal M$ can be computed as
$$\mathcal G'_{\eta, i}(x^*) = \mathds 1_D - \eta \operatorname{Hess} \mathcal L_i (x^*) =  \mathds 1_D - \eta\nabla_x\mathfrak F(x^*,y_i)\nabla_x\mathfrak F(x^*,y_i)^t.$$
For points $x \in \mathbb R^D$ close to $x^*$, we have
$$\big[\mathcal G_{\eta, \xi_n} \circ \dots \circ \mathcal G_{\eta, \xi_1}\big] (x) = x^* + \big[\mathcal G'_{\eta, \xi_n}(x^*) \dots \mathcal G'_{\eta, \xi_1}(x^*)\big] (x-x^*) + o(\|x-x^*\|).$$
Note that the matrices $\mathcal G'_{\eta, i}(x^*)$ also all respect the splitting $\mathbb R^D = \mathcal T(x^*) \oplus \mathcal N(x^*)$ with $\mathcal G'_{\eta, i}(x^*)|_{\mathcal T(x^*)} = \operatorname{Id}_{\mathcal T(x^*)}$.
Analogously to gradient descent, we would like to call $x^*$ linearly stable if
$$\lambda(x^*) = \lim_{n \to \infty} \frac{1}{n}\log\big(\big\|\mathcal G'_{\eta, \xi_n}(x^*) \dots \mathcal G'_{\eta, \xi_1}(x^*)|_{\mathcal N(x^*)}\big\|\big) <0.$$
At first glance, this definition raises two problems. First, it is not clear whether the limit exists. Secondly, the value of $\lambda(x^*)$ seems to depend on the random sequence $(\xi_1, \xi_2, \dots)$. 
However, Kingman's sub-additive ergodic theorem (\citealp{Kingman}, see also \citealp{SteeleKET})\footnote{The integrability condition for Kingman's ergodic theorem holds trivially here, as we only consider finitely many different matrices.} states that for almost every realization $(\xi_1, \xi_2, \dots)$ we have
$$\lim_{n \to \infty} \frac{1}{n}\log\big(\big\|\mathcal G'_{\eta, \xi_n}(x^*) \dots \mathcal G'_{\eta, \xi_1}(x^*)|_{\mathcal N(x^*)}\big\|\big) = \inf_{n \in \mathbb N} \frac{1}{n} \mathbb E\big[\log\big(\big\|\mathcal G'_{\eta, \xi_n}(x^*) \dots \mathcal G'_{\eta, \xi_1}(x^*)|_{\mathcal N(x^*)}\big\|\big)\big].$$
Hence, using the right-hand side as a definition for $\lambda(x^*)$, i.e.
\begin{equation}\label{eq:lambdaDef}
    \lambda(x^*) := \inf_{n \in \mathbb N} \frac{1}{n} \mathbb E\big[\log\big(\big\|\mathcal G'_{\eta, \xi_n}(x^*) \dots \mathcal G'_{\eta, \xi_1}(x^*)|_{\mathcal N(x^*)}\big\|\big)\big] \in [-\infty, \infty),
\end{equation}
solves both issues. It should be noted that the sequence 
$$n \mapsto \frac{1}{n} \mathbb E\big[\log\big\|\mathcal G'_{\eta, \xi_n}(x^*) \dots \mathcal G'_{\eta, \xi_1}(x^*)|_{\mathcal N(x^*)}\big\|\big]$$
is monotonically decreasing and one may equivalently define $\lambda(x^*)$ by
\begin{equation}\label{eq:lambdaAsLim}
    \lambda(x^*) := \lim_{n \to \infty} \frac{1}{n} \mathbb E\big[\log\big\|\mathcal G'_{\eta, \xi_n}(x^*) \dots \mathcal G'_{\eta, \xi_1}(x^*)|_{\mathcal N(x^*)}\big\|\big] \in [-\infty, \infty).
\end{equation}
In fact, the following theorem due to \cite{Osdeledec68} (see also e.g.~chapter 3.4 in \citealp{Bible}) shows the existence of so-called \emph{Oseledec's subspaces}, which are the equivalent notion to generalized eigenspaces for random-matrix products.
\begin{theorem}[Multiplicative Ergodic Theorem]\label{theo:Oseledets}
    There exists a natural number $1\leq k \leq N$ and a tuple of extended real numbers
    $$-\infty \leq \lambda_k < \dots < \lambda_2 < \lambda_1 = \lambda(x^*)$$
and a tuple of multiplicities $m_1, m_2, \dots, m_k$ with $m_1 + \dots + m_k = N$ such that the following holds:
For almost every\footnote{with respect to the uniform i.i.d.~measure,} sequence $(\xi_1, \xi_2, \dots)$ there exists a tuple of subspaces
$$\{0\} = V_{k+1}(\xi_1, \xi_2, \dots) \subsetneq V_k(\xi_1, \xi_2, \dots) \subsetneq \dots \subsetneq V_2(\xi_1, \xi_2, \dots)\subsetneq V_1(\xi_1, \xi_2, \dots) = \mathcal N(x^*)$$
with $\operatorname{dim}(V_i) = m_k + \dots + m_i$ such that for every $1 \leq i \leq k$
$$\lim_{n \to \infty} \frac{1}{n} \log \left[ \big\|\mathcal G'_{\eta, \xi_n}(x^*) \dots \mathcal G'_{\eta, \xi_1}(x^*)y \big\|\right] = \lambda_i, ~ \forall\, y \in V_i(\xi_1, \xi_2, \dots) \setminus V_{i+1}(\xi_1, \xi_2, \dots).$$
In particular, for every $y \in \mathcal N \setminus V_2(\xi_1, \xi_2, \dots)$, we have
$$\lim_{n \to \infty} \frac{1}{n} \log \left[ \big\|\mathcal G'_{\eta, \xi_n}(x^*) \dots \mathcal G'_{\eta, \xi_1}(x^*)y \big\|\right] = \lambda(x^*).$$
\end{theorem}
We will call $x^*$ \emph{linearly stable} (or \emph{linearly unstable}) under stochastic gradient descent if $\lambda(x^*)< 0$ (or $\lambda(x^*) > 0$, respectively).\footnote{As mentioned in the introduction, this definition differs from the ones introduced by \cite{WuMaE} and \cite{maYing}. See Appendix \ref{app:WuMaW} for a detailed comparison.}

\subsection{Main Result}\label{sec:mainRes}
The main contribution of our work is to rigorously show that the definitions for linear stability/instability, heuristically derived in Section \ref{sec:linstab}, characterize the supports of $X^{\operatorname{GD/SGD}}_{\lim}$ and thus characterize the qualitative implicit bias of gradient descent/stochastic gradient descent. 
In our derivation, we made some assumptions which can be formalized as follows.
\begin{hypothesis}\label{assu:manifold}
    For each $x \in \mathcal M$ the set of vectors
    $$\left\{\nabla_x \mathfrak F(x,y_i) : i \in [N]\right\} \subset R^D$$
    is linearly independent. In particular, $\mathcal M$ is an embedded submanifold of $\mathbb R^D$.
\end{hypothesis}
\begin{hypothesis}\label{assu:init}
    The initial distribution $\nu$ is equivalent to the Lebesgue measure, i.e.
    $$\nu(B) = 0 \Leftrightarrow \operatorname{Leb}(B) = 0, ~\forall \, B \in \mathfrak B(\mathbb R^D).$$
\end{hypothesis}
\noindent Recall that, by an argument by \cite{YCooper}, Hypothesis \ref{assu:manifold} is fulfilled for generic training data, while \ref{assu:init} holds for the most common choice of initial distribution. 

Furthermore, the heuristic argument presented in the previous section relied on the assumption that with probability one the optimization algorithms do not reach $\mathcal M$ in a finite number of steps. Hypothesis \ref{assu:init} ensures that, with probability one, the algorithms are not initialized on $\mathcal M$. However, the setting presented so far is too general to exclude that the optimization algorithms reach $\mathcal M$ after any positive finite number of iterations with positive probability. Thus, we require an additional assumption.
From now on, consider a fixed learning rate $\eta>0$.
\begin{hypothesis}\label{assu:nonsing}
The map $\mathcal G_\eta$ and the maps $\mathcal G_{\eta, 1}, \dots, \mathcal G_{\eta, N}$, defined in \eqref{eq:Geta} and \eqref{eq:Getai}, are non-singular, i.e.~the pre-image of every Lebesgue-null set is a Lebesgue-null set. 
\end{hypothesis}
\noindent In particular, a continuously differentiable map $\mathcal G: \mathbb R^D \to \mathbb R^D$ is non-singular if its Jacobian is invertible Lebesgue almost everywhere. Whether this is true for the maps $\mathcal G_\eta, \mathcal G_{\eta, 1}, \dots, \mathcal G_{\eta, N}$ depends on the network function $\mathfrak F$. Yet, it is reasonable to assume that this should be satisfied for the common neural network architectures, at least for almost every learning rate $\eta$.

Recall that the support of $X^{\operatorname{GD/SGD}}_{\lim}$ is defined as
$$\operatorname{supp} \left(X^{\operatorname{GD/SGD}}_{\lim}\right) := \left\{x \in \mathcal M :U \subseteq\mathcal M \text{ open nbhd. of } x \Rightarrow \mathbb P\left(X^{\operatorname{GD/SGD}}_{\lim} \in U\right) >0 \right\}.$$
For gradient descent, we will show the following main result.
\begin{theoremMain}\label{theo:mainGD}
Suppose \ref{assu:manifold}, \ref{assu:init}, and \ref{assu:nonsing} are satisfied. Let $x^* \in \mathcal M$.
\begin{enumerate}
    \item[(i)] If $\mu(x^*) < 0$, then $x^* \in\operatorname{supp} \left(X^{\operatorname{GD}}_{\lim}\right)$.
    \item[(ii)] If $\mu(x^*) > 0$, then $x^* \notin\operatorname{supp} \left(X^{\operatorname{GD}}_{\lim}\right)$. 
\end{enumerate}
\end{theoremMain}
The analogous result for stochastic gradient descent requires some additional assumptions on the global minimum $x^*$ in question.
\begin{definition}\label{def:regular}
    A global minimum $x^* \in \mathcal M$ is said to be regular if
    \begin{enumerate}
        \item[(i)] for every $i \in [N]$, we have
        \begin{equation}\label{etacond}
           \|\nabla_x\mathfrak F(x^*,y_i)\|^2 \notin \left\{\frac{1}{\eta}, \frac{2}{\eta}\right\}, \text{ and}
        \end{equation}
        \item[(ii)] there exists no proper sub-set $\emptyset \subsetneq \mathcal A \subsetneq [N]$, such that 
        $$\nabla_x\mathfrak F(x^*,y_i) \cdot \nabla_x\mathfrak F(x^*,y_j) = 0,~\forall \, i \in \mathcal A,\, j \in [N] \setminus \mathcal A.$$
    \end{enumerate}
\end{definition}
\noindent Note that almost every family of vectors $(\nabla_x\mathfrak F(x^*,y_i))_{i \in [N]}$ will satisfy these conditions. Thus, it is reasonable to assume that, for most network functions $\mathfrak F$, almost every global minimum $x^* \in \mathcal M$ is regular.
\revision{Together, the conditions (i) and (ii) in Definition \ref{def:regular} imply that the semigroup generated by the operators $\mathcal G'_{\eta, 1}(x^*), \dots, \mathcal G'_{\eta, N}(x^*)$ has nice algebraic properties, namely being contracting and strongly irreducible (cf.~Definition \ref{def:csi} below). These properties are well established in the literature on random-matrix products (see, e.g.~\citealp{RMP}) and allow us to use tools from that theory which are crucial for our proof.} 
With this definition in hand, we get the following result for SGD.
\begin{theoremMain}\label{theo:mainSGD}
Suppose \ref{assu:manifold}, \ref{assu:init}, and \ref{assu:nonsing} are satisfied. Let $x^* \in \mathcal M$ be regular.
\begin{enumerate}
    \item[(i)] If $\lambda(x^*) < 0$, then $x^* \in\operatorname{supp} \left(X^{\operatorname{SGD}}_{\lim}\right)$.
    \item[(ii)] If $\lambda(x^*) > 0$, then $x^* \notin\operatorname{supp} \left(X^{\operatorname{SGD}}_{\lim}\right)$. 
\end{enumerate}
\end{theoremMain}

\begin{remark}
    Both Theorem \ref{theo:mainGD} and Theorem \ref{theo:mainSGD} do not address the case $\mu(x^*) = 0$, respectively $\lambda(x^*) = 0$. Since the support of $X_{\lim}^{GD/SGD}$ is closed by definition and $\mu: \mathcal M \to \mathbb R$ is continuous, most points with $\mu(x^*) = 0$ should have global minima $x' \in \mathcal M$ with $\mu(x') < 0$ near them and therefore be in the support of $X_{\lim}^{GD}$. Making this argument rigorous would require a more precise knowledge of the network function $\mathfrak F$ and is beyond the scope of this paper. The continuity of $\lambda$ is a more subtle issue. In general, $\lambda: \mathcal M \to \mathbb R$ is only upper semi-continuous\footnote{This can readily be seen from the fact that it is defined as an infimum of continuous function.}. However, a recent result of \cite{AvilaEskenViana} shows that $\lambda$ is continuous at all points $x^*$, where the matrices $\mathcal G'_{\eta, 1}(x^*),\dots, \mathcal G'_{\eta, N}(x^*)$ are all invertible. This is, in particular, the case for all regular $x^* \in \mathcal M$.
\end{remark}

\subsection{Relation to the Neural Tangent Kernel}\label{sec:ntk}
We point out that, for any $x^*\in \mathcal M$, the values of $\mu(x^*)$ and $\lambda(x^*)$ can be deduced entirely from the \emph{neural tangent kernel} $K_{x^*}: \mathbb R^d \times \mathbb R^d \to \mathbb R$, first introduced by \cite{NTK}, which is defined by
$$K_{x^*}(y,y') = \nabla_x\mathfrak F(x^*, y)^t\nabla_x\mathfrak F(x^*, y'),$$
or more specifically, its \emph{Gram matrix} $G_{x^*} \in \mathbb R^{N\times N}$ given by
$$[G_{x^*}]_{i,j} := K_{x^*}(y_i,y_j).$$
In more detail, let $S_{x^*} \in \mathbb R^{D \times N}$ denote the matrix whose $i$-th column is given by $\nabla_x \mathfrak F(x^*, y_i)$, i.e.
\begin{equation}\label{eq:SDef}
    S_{x^*} := \begin{pmatrix}|&&|\\\nabla_x \mathfrak F(x^*, y_1)& \dots& \nabla_x \mathfrak F(x^*, y_N)\\|&&|\end{pmatrix}.
\end{equation}
Clearly, $S_{x^*}$ maps $\mathbb R^N$ isomorphically onto $\mathcal N(x^*)$ and straightforward calculations, using linear independence of the gradients, show that 
\begin{align*}
    [\operatorname{Hess} \mathcal L (x^*)] S_{x^*} &= \frac{1}{N}S_{x^*}G_{x^*}&&\text{and}& [\operatorname{Hess} \mathcal L_i (x^*)] S_{x^*} &= S_{x^*}G_{x^*,[i]},
\end{align*}
where $G_{x^*,[i]}$ denotes the matrix $G_{x^*}$ with every row but the $i$-th set to zero. It should be noted that $S_{x^*}$ is not a square matrix and is thus not invertible as a matrix. However, as an isomorphism from $\mathbb R^N$ to $\mathcal N(x^*)$ it can be inverted, and we let $S_{x^*}^{-1} \in \mathbb R^{N\times D}$ denote the matrix associated with this inverse isomorphism. It satisfies $S^{-1}_{x^*}S_{x^*} = \mathds 1_N$ and $(S_{x^*}S^{-1}_{x^*})|_{\mathcal N(x^*)} = \mathds 1_D|_{\mathcal N(x^*)}$. Thus, the restrictions of the Jacobians of $\mathcal G_\eta$ and $\mathcal G_{\eta, i}$ to $\mathcal N(x^*)$ can be expressed as
\begin{equation}\label{eq:conjGD}
    \mathcal G'_\eta(x^*)|_{\mathcal N(x^*)} = S_{x^*}\left(\mathds 1_N -\frac{\eta }{N}G_{x^*}\right){S_{x^*}}^{-1}
\end{equation}
and 
\begin{equation}\label{eq:conjSGD}
    \mathcal G'_{\eta,i}(x^*)|_{\mathcal N(x^*)} = S_{x^*}\left(\mathds 1_N -\eta G_{x^*,[i]}\right){S_{x^*}}^{-1}.
\end{equation}
From this we get $\rho_{\operatorname{Spec}}(\mathcal G'_\eta(x^*)|_{\mathcal N(x^*)}) = \rho_{\operatorname{Spec}}(\mathds 1_N -\frac{\eta }{N}G_{x^*})$ and, using the symmetry of $(\mathds 1_N -\frac{\eta }{N}G_{x^*})$, we also get 
\begin{equation}\label{eq:NTKmuDef}
    \mu(x^*) = \log\left[\rho_{\operatorname{Spec}}\left(\mathds 1_N -\frac{\eta }{N}G_{x^*}\right)\right] = \log \left\|\mathds 1_N -\frac{\eta }{N}G_{x^*}\right\|.
\end{equation}
An analogous statement can be obtained for $\lambda(x^*)$. Indeed, we have the estimates
\begin{align*}
    &~~~\,\log\big\|\left(\mathcal G'_{\eta, \xi_n}(x^*) \dots \mathcal G'_{\eta, \xi_1}(x^*)\right)|_{\mathcal N(x^*)}\big\| \\
    &= \log\big\|S_{x^*}\left(\mathds 1_N -\eta G_{x^*,[\xi_n]}\right)\dots \left(\mathds 1_N -\eta G_{x^*,[\xi_1]}\right){S_{x^*}}^{-1}\big\|\\
    &\leq \log\|S_{x^*}\|+\log\big\|\left(\mathds 1_N -\eta G_{x^*,[\xi_n]}\right)\dots \left(\mathds 1_N -\eta G_{x^*,[\xi_1]}\right)\big\| + \log \left\|{S_{x^*}}^{-1}\right\|\\
\end{align*}
and 
\begin{align*}
    &~~~\,\log\big\|\left(\mathds 1_N -\eta G_{x^*,[\xi_n]}\right)\dots \left(\mathds 1_N -\eta G_{x^*,[\xi_1]}\right)\big\|\\
    &=\log\big\|{S_{x^*}}^{-1}\left(\mathcal G'_{\eta, \xi_n}(x^*) \dots \mathcal G'_{\eta, \xi_1}(x^*)\right)S_{x^*}\big\|\\
    &\leq \log \left\|{S_{x^*}}^{-1}\right\|+\log\big\|\left(\mathcal G'_{\eta, \xi_n}(x^*) \dots \mathcal G'_{\eta, \xi_1}(x^*)\right)|_{\mathcal N(x^*)}\big\| + \log\|S_{x^*}\|.\\
\end{align*}
Together, they imply 
\begin{align}
    \lambda(x^*) &= \lim_{n \to \infty} \frac{1}{n} \mathbb E\big[\log\big\|\mathcal G'_{\eta, \xi_n}(x^*) \dots \mathcal G'_{\eta, \xi_1}(x^*)|_{\mathcal N(x^*)}\big\|\big]\nonumber\\
    &=\lim_{n \to \infty} \frac{1}{n} \mathbb E\big[\log\big\|\left(\mathds 1_N -\eta G_{x^*,[\xi_n]}\right)\dots \left(\mathds 1_N -\eta G_{x^*,[\xi_1]}\right)  \big\|\big]\nonumber\\
    &=\inf_{n \in \mathbb N} \frac{1}{n} \mathbb E\big[\log\big\|\left(\mathds 1_N -\eta G_{x^*,[\xi_n]}\right)\dots \left(\mathds 1_N -\eta G_{x^*,[\xi_1]}\right)  \big\|\big]\label{eq:NTKlambdaDef}.
\end{align}

This observation is useful for two reasons. On the one hand, it connects the central quantities in the present work to an ongoing line of research of neural networks in the infinite-width limit (see \citealp{NTK}). Additionally, it turns out that the expressions \eqref{eq:NTKmuDef} and \eqref{eq:NTKlambdaDef} are nicer to work with than the corresponding expressions in the previous section and will be used in the proofs of the main theorems. \revision{Finally, these expressions are well defined for any $x \in \mathbb R^D$ and not only for global minima $x^* \in \mathcal M$. This is especially useful for the potential empirical studies suggested in the following section.}

\revision{\subsection{Edge of Stability and Empirical Studies}\label{sec:EoS}
While the present work is purely theoretical, the Lyapunov exponent $\lambda(x^*)$ of a global minimum $x^*$ is also an interesting object for empirical studies. As mentioned in the introduction, a highly interesting study by \cite{Cohen22} observed that training with gradient descent undergoes two distinct phases. At initialization, the Hessian of the loss landscape satisfies $\|\operatorname{Hess} \mathcal L(X_0^{\operatorname{GD}})\| < \frac{2}{\eta}$. During the early stages of training with gradient descent, the parameters $X_n^{\operatorname{GD}}$ move to progressively sharper regions of the loss landscape, that is, the operator norm of the Hessian of the loss function increases. During this phase, called \emph{progressive sharpening}, the loss $\mathcal L(X_n^{\operatorname{GD}})$ decreases monotonically.
Once $\|\operatorname{Hess} \mathcal L(X_n^{\operatorname{GD}})\|$ reaches the threshold $\frac{2}{\eta}$, progressive sharpening stops and training continues in the so-called \emph{edge-of-stability} phase. During this phase, the Hessian of the loss function remains approximately constant close to the stability threshold $\frac{2}{\eta}$. Meanwhile, the loss is still decreasing in the long run, but no longer monotonically. 
\cite{Cohen22} also studied whether stochastic gradient descent exhibits the same behavior. In Appendix G, they observe that for stochastic gradient descent progressive sharpening stops before the threshold $\|\operatorname{Hess} \mathcal L(X_n^{\operatorname{GD}})\| \approx \frac{2}{\eta}$, is reached, suggesting that SGD does not enter the edge of stability. In Appendix H, however, they observe that the expected change of the loss does behave similarly to the non-monotonous decrease observed for gradient descent at the edge of stability. Based on this, they conjecture (cf. Section 6 in \citealp{Cohen22}) that SGD does enter an edge-of-stability regime, except that $\|\operatorname{Hess} \mathcal L(x)\| \leq \frac{2}{\eta}$ is no longer the correct notion of stability for gradient descent (cf.~also the discussions in \citealp{WuMaE} and \citealp{andreyev2025}).

Instead of directly tracking the sharpness of the loss, $\|\operatorname{Hess} \mathcal L(x)\|$, one can equivalently study the evolution of $\mu(x)$ along the trajectory. It should be mentioned that in Section \ref{sec:linstab} we have only defined $\mu(x)$ for the parameters $x$ that lie in the manifold of interpolation solutions $\mathcal M$. However, the equivalent definition \eqref{eq:NTKmuDef} of $\mu$ given in Section \ref{sec:ntk}, is well defined for any $x \in \mathbb R^D$. Thus, it is possible to track $\mu(X_n^{\operatorname{GD}})$ during training. Since $\mu$ is related to the Hessian of the loss by $\mu(x) = \log\big|1-\eta \|\operatorname{Hess} \mathcal L(x)\|\big|$, it contains the same information as $\|\operatorname{Hess} \mathcal L(x)\|$. In the experiments of \cite{Cohen22}, we would observe that $\mu(X_0^{\operatorname{GD}})$ is negative and that $\mu(X_n^{\operatorname{GD}})$ progressively increases until it enters the edge of stability $\mu(X_n^{\operatorname{GD}}) = 0$ from which point on $\mu(X_n^{\operatorname{GD}})$ remains close to zero. Based on our main results, we suggest that from a dynamical point of view, tracking $\lambda(X_n^{\operatorname{SGD}})$ during training with stochastic gradient descent is the appropriate analogue to the study of \cite{Cohen22}. 
We leave it as an open problem to the community to determine the behavior of $\lambda$ by the means of empirical experiments.
An interesting challenge in conducting such a study lies in the numerical computation of the Lyapunov exponent for high-dimensional random-matrix products. Numerical schemes for the computation of Lyapunov exponents can be found, for example, in \cite{RuelleEckmann85} and \cite{sandri1996numerical}.
}

\subsection{Possible Extensions and Outlook}\label{sec:open}
While our main results are general in the sense that they require only very weak assumptions on the network function $\mathfrak F$, we want to point out that we only consider scalar regression problems and training with gradient descent or stochastic gradient descent with mini-batch size 1. As mentioned above, we have made these restrictions to simplify the already extensive proofs as much as possible. The main results can be extended, with minor modifications, to a more general setting.

In practice, SGD is usually implemented with so-called mini-batches to allow for parallel computations. For SGD with mini-batches of size $1 \leq B \leq N$, the iterative step \eqref{eq:SGD} is replaced by 
$$X_{n+1}^{\operatorname{SGD}} := X_n^{\operatorname{SGD}} - \eta\nabla \mathcal L_{\Xi_{n+1}}(X_n^{\operatorname{SGD}}),$$
where $(\Xi_n)_{n \in\mathbb N}$ is an i.i.d. sequence of size $B$ subsets $\Xi_n \subseteq [N]$ chosen uniformly from the $\begin{pmatrix}N\\B\end{pmatrix}$ possible subsets and 
$$\mathcal L_{\Xi}(x) := \frac{1}{B} \sum_{i \in \Xi} \mathcal L_i(x).$$
For $B=1$ this coincides with the SGD algorithm presented in Section \ref{subs:sgd} and for $B = N$ the iterative step is deterministic and coincides with GD. Stochastic gradient descent with mini-batch size $1<B<N$ can thus be seen as an interpolation between the two algorithms.
The derivations from sections \ref{sec:linstab} and \ref{sec:ntk} can be carried out analogously for mini-batch SGD and one can express the Lyapunov exponent of mini-batch SGD with learning rate $\eta$ and mini-batch size $B$ as
$$\lambda(x^*) = \inf_{n \in \mathbb N} \frac{1}{n} \mathbb E\left[\log\left\|\left(\mathds 1_N -\frac{\eta}{B} G_{x^*,[\Xi_n]}\right)\dots \left(\mathds 1_N -\frac{\eta}{B} G_{x^*,[\Xi_1]}\right)  \right\|\right],$$
where $G_{x^*,[\Xi]}$ denotes the Gram matrix $G_{x^*}$ of the neural tangent kernel with all rows whose index is not in $\Xi$ set to 0. In order to derive a version of Theorem \ref{theo:mainSGD} it is necessary to adapt the notion of regular global minima (cf.~Definition \ref{def:regular}) to ensure that the analog of Lemma \ref{lemm:CSI} in the proof below still remains valid. We leave this as a problem for future work.
The rest of the proof of Theorem \ref{theo:mainSGD} can be extended to mini-batch SGD with only minor changes. Such an extension of Theorem \ref{theo:mainSGD} could help to better understand the impact of the learning rate and the mini-batch size on generalization properties (see e.g.~\citealp{Hofferetal17,GoyalEtAl,keskar2017b} for numerical studies).

Furthermore, it would be interesting to extend our analysis to other \revision{optimizers} such as SGD with momentum (cf.~\citealp{Rumelhart86}) or Adam (cf.~\citealp{Kingma14}). The derivations in Section \ref{sec:linstab} can be extended to these algorithms by linearizing the iteration steps around fixed points where all moment terms are zero. In the case of Adam, the parameter $\epsilon$ appearing in the numerator of the final update step is usually chosen extremely small (\citealp{Kingma14} suggest $\epsilon = 10^{-8}$). Thus, the linearization only approximates the dynamics of the actual algorithm in a vanishingly small neighborhood, and it is questionable whether it is still meaningful for the dynamics of the algorithm.

Our analysis can be extended to multidimensional regression problems with ground truth function $\hat f: \mathbb R^d \to \mathbb R^{\tilde d}$ and network function $\mathfrak F: \mathbb R^D \times \mathbb R^d \to \mathbb R^{\tilde d}$. In fact, one can interpret such a multidimensional regression problem as a scalar regression problem with a ground truth function $\hat f: \mathbb R^d \times [\tilde d] \to \mathbb R$ and a network function $\mathfrak F: \mathbb R^D \times (\mathbb R^d \times [\tilde d]) \to \mathbb R$. This has the effect of splitting one training example $(y,z) \in \mathbb R^d \times \mathbb R^{\tilde d}$ into $\tilde d$ training examples $((y,1),z_1), \dots, ((y,\tilde d), z_{\tilde d}) \in (\mathbb R^d \times [\tilde d]) \times \mathbb R$. An SGD step with the training example $(y,z)$ now corresponds to an SGD step with the mini-batch $((y,1),z_1), \dots, ((y,\tilde d), z_{\tilde d})$. Similarly to mini-batch SGD, the main challenge in extending Theorem \ref{theo:mainSGD} to multidimensional regression problems is to adapt the notion of a regular global minimum (cf.~Definition \ref{def:regular}) in such a way that Lemma \ref{lemm:CSI} still remains valid for multidimensional regression.

Unfortunately, our results cannot be easily extended to classification problems. For a classification task with $k$ classes, it is common to encode the training data as pairs $(y,z) \in \mathbb R^d \times \mathbb R^k$, where $z = e_i$ is the $i$-th unit vector, where $i \in [k]$ is the class $y$ belongs to. Then one considers a network function $\mathfrak F: \mathbb R^D \times \mathbb R^d \to \mathbb R^k$, which is then postcomposed with the softmax function $\operatorname{softmax}_\beta: \mathbb R^k \to [0,1]^k$ given by
$$\operatorname{softmax}_\beta(z_1,\dots z_k)_i := \frac{e^{\beta z_i}}{\sum_{j =1}^k e^{\beta z_j}},$$
where $\beta>0$ is some positive real number commonly referred to as inverse temperature. For training, the most common loss function is the cross-entropy loss given by
$$\ell(\bar z, z) := -\sum_{i = 1}^n z_i \log (\bar z_i).$$
To have $\ell(\bar z, z) = 0$, one necessarily needs $\bar z = z$. However, the outputs of the softmax function are in $(0,1)^k$ for finite input values. Thus, it is not possible to reach training error zero and $\mathcal M = \emptyset$.

Finally, it should be mentioned that our analysis assumes that the model is trained to convergence. In practice, it has been observed that stopping the algorithm early can improve generalization. While the global minima with $\lambda(x^*)>0$ are asymptotically unstable, the finite time Lyapunov exponents
$$\lambda_{w,n}(x^*) := \frac{1}{n}\log \left\|\left(\mathds 1_N -\eta G_{x^*,[\xi_n]}\right)\dots \left(\mathds 1_N -\eta G_{x^*,[\xi_1]}\right)w\right\|$$
can, even for large $n$, still be negative with small, but positive probability. This effect is captured by a large deviation principle (cf.~\citealp{ArnoldKliemann87}), the rate function of which depends on the moment Lyapunov exponents (see Definition \ref{def:MLE}).

\section{Proofs of the Main Results}\label{sec:proofs}
\subsection{Overview}\label{sec:proofOverview}
The remainder of this paper will consist of the proofs of Theorem \ref{theo:mainGD} and Theorem \ref{theo:mainSGD}. For the main arguments, it will be convenient to work in a local coordinate system in which $\mathcal M$ corresponds to a linear subspace. Such a coordinate system will be introduced in Section \ref{sec:localco}. Theorem \ref{theo:mainGD} (i) then follows from an elementary argument presented in Section \ref{sec:gdstable}. In order to prove Theorem \ref{theo:mainGD} (ii), in Section \ref{sec:gdcenterstable}, we employ a center-stable manifold theorem. In Section \ref{sec:RDS}, we introduce a random-dynamical-system framework to treat stochastic gradient descent. Theorem \ref{theo:mainSGD} (i) is then proved in Section \ref{sec:sgdstable}. The proof is similar to the proof of Theorem \ref{theo:mainGD} (i) presented in Section \ref{sec:gdstable}, but the possibility of non-uniform hyperbolicity adds an additional challenge. Finally, Theorem \ref{theo:mainSGD} (ii) is proved in sections \ref{sec:semiGr}-\ref{sec:sgdunstable}. The proof is inspired by previous work on the instability of invariant subspaces for stochastic differential equations 
(see, e.g., \citealp{BaxendaleStroock, Baxendale1991}). While these works rely on Hörmander conditions (cf.~\citealp{Hormander}), we use a criterion by \cite{lepage} to establish a spectral gap for the projective semigroup. The assumptions for the criterion of Le Page are checked in Section \ref{sec:semiGr}. In Section \ref{sec:LyapFun}, we construct a local Lyapunov function similar to the recent works of \cite{BedrossianBlumenthalPunshon2022} and \cite{BlumenthalCotiGvalani2023}. The proof of Theorem \ref{theo:mainSGD} is then completed in Section \ref{sec:sgdunstable}.

Throughout the entire Section \ref{sec:proofs} we assume \ref{assu:manifold}, \ref{assu:init} and \ref{assu:nonsing} as standing assumptions.

\subsection{Local Coordinates}\label{sec:localco}
In the following, for some fixed $x^* \in \mathcal M$, we introduce a local coordinate system for a neighborhood of $x^*$, in which the generally nonlinear manifold $\mathcal M$ becomes a linear subspace aligned with the coordinate axes. Using these coordinates will be helpful in all further proofs.
\begin{lemma}\label{lemm:localco}
There exist an open neighborhood $x^* \in  \hat U\subset \mathbb R^D$ and an open neighborhood $(0,0) \in \hat V \subseteq \mathbb R^{D-N} \times \mathbb R^N$, as well as a smooth diffeomorphism $\chi: \hat V \to \hat U$, such that 
\begin{enumerate}
    \item[(i)] $\chi(0,0) = x^*$,
    \item[(ii)] $\chi( \hat V \cup (\mathbb R^{D-N} \times \{0\})) = \hat U \cap \mathcal M,$
    \item[(iii)] the Jacobian at the origin is given by 
    \begin{equation}\label{eq:chiJacobi}
        \chi'(0,0) = \begin{pmatrix}A &S_{x^*}\end{pmatrix},
    \end{equation}
    where $A\in \mathbb R^{D \times (D-N)}$ is some matrix which induces an isomorphism from $\mathbb R^{D-N}$ onto $\mathcal T(x^*)$ and $S_{x^*} \in \mathbb R^{D \times N}$ is the matrix defined in \eqref{eq:SDef},
    \item[(iv)] and $\chi$ is bi-Lipschitz with Lipschitz constant $L_\chi$, i.e.~both $\chi$ and $\chi^{-1}$ are Lipschitz continuous with said Lipschitz constant.
\end{enumerate}
\end{lemma}

\begin{proof}
    Let $A\in \mathbb R^{D \times (D-N)}$ be some matrix that induces an isomorphism from $\mathbb R^{D-N}$ to $\mathcal T(x^*)$.
    Since $\mathcal M$ is an embedded smooth manifold, we can find neighborhoods $0 \in U_\mathcal T \subseteq \mathcal T(x^*)$ and $x^* \in \tilde U \subseteq \mathbb R^D$ and a smooth map $\zeta: U_\mathcal T \to \mathcal N(x^*)$ with $\zeta(0) = 0$, $\zeta'(0) = 0$ and 
    $$\{x^*+x+\zeta(x) : x \in U_\mathcal T\} = \tilde U \cap \mathcal M.$$
    This allows us to define a smooth map $\chi: A^{-1}U_\mathcal T \times \mathbb R^N \to \mathbb R^D$ by
    $$\chi(v,w) = x^*+Av+S_{x^*}w+\zeta(Av).$$
    It can be easily verified that $\chi$ is injective and 
    $$\chi(A^{-1}U_\mathcal T \times \{0\}) = \tilde U \cap \mathcal M.$$ 
    Also, the Jacobian is given by
    $$\chi'(v,w)  = \begin{pmatrix}A+\zeta'(Av)A& S_{(x_0)}\end{pmatrix},$$
    so in particular $\chi$ is a local diffeomorphism and \eqref{eq:chiJacobi} holds.
    Set $\hat V := \chi^{-1}(\tilde U)$ and $\hat U:= \chi(\hat V)$.
    If we now let $\chi$ be its restriction $\chi: \hat V \to \hat U$, it is a smooth diffeomorphism, satisfying i) to iii). If $\chi$ is not bi-Lipschitz, we can reassign $\hat V$ to a smaller neighborhood, which is precompact in the original $\hat V$. After reassigning $\hat U$ and restricting $\chi$ accordingly, $\chi$ will be bi-Lipschitz.
\end{proof}
\begin{figure}
    \centering
    \begin{overpic}[scale=.25,,tics=10, width = \textwidth]
{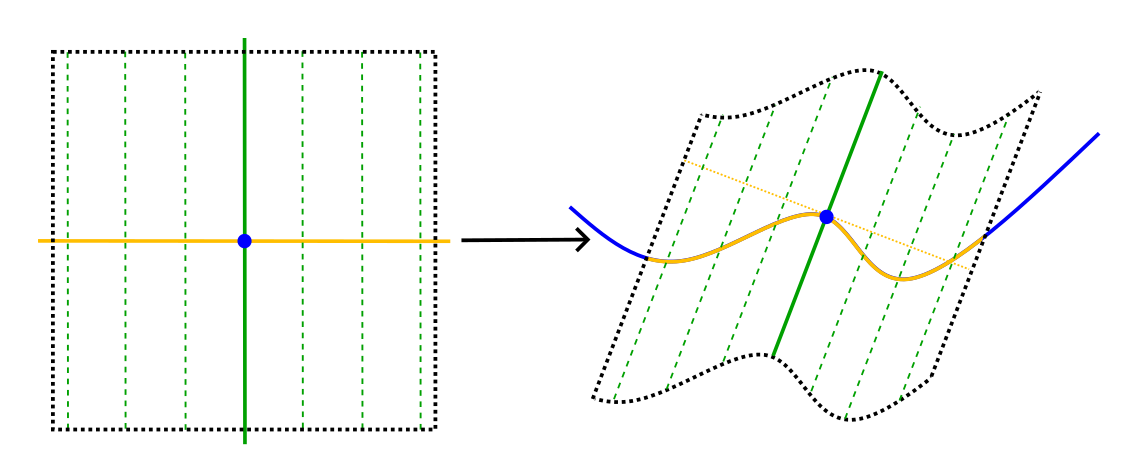}
\put(45,23){$\chi$}
\put(23,39){$\hat V$}
\put(81,33){$\hat U$}
\put(92,23){\color{blue} $\mathcal M$}
\put(23,22){\color{blue} $(0,0)$}
\put(74,23){\color{blue} $x^*$}
\put(5,16){\color{orange} $\mathbb R^{D-N} \times \{0\}$}
\put(57,15){\color{orange} $\hat U\cap \mathcal M$}
\put(22,8){\color{Green}$\{0\} \times\mathbb R^N$}
\put(70,12){\color{Green}$\hat U \cap (x^*+\mathcal N(x^*))$}
\end{overpic}
    \caption{Schematic representation of the construction of $\chi$. Objects with the same color are mapped onto each other.}
    \label{fig:loccord}
\end{figure}
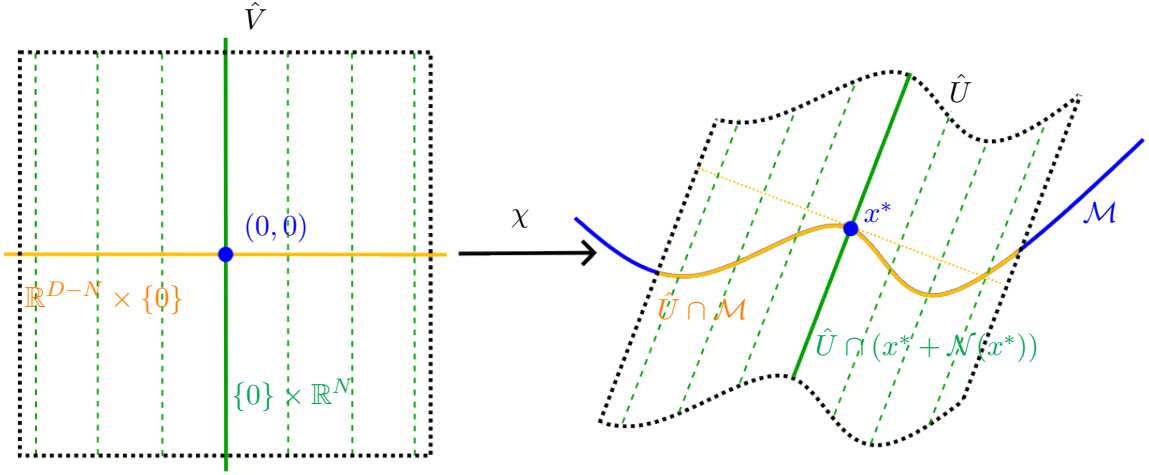

The learning dynamics of (stochastic) gradient decent can, at least locally, be lifted in the new coordinates via the function $\chi$. We define the functions $\hat{\mathcal G}_{\eta}, \hat{\mathcal G}_{\eta,1}, \dots \hat{\mathcal G}_{\eta, N}: V^* \to \hat V$ by 
\begin{align*}
    \hat{\mathcal G}_{\eta}(v,w) &= \chi^{-1}(\mathcal G_{\eta}(\chi(v,w)))&&\text{and} &\hat{\mathcal G}_{\eta,i}(v,w) &= \chi^{-1}(\mathcal G_{\eta,i}(\chi(v,w))),
\end{align*}
where $V^*\subseteq \hat V$ is an open set given by
\begin{equation}\label{eq:ustar}
    V^* :=  \chi^{-1}\left(\mathcal G_\eta^{-1}(\hat U) \cap \bigcap_{i \in [N]}\mathcal G_{\eta,i}^{-1}(\hat U)\right).
\end{equation}
Let $\tau: V^* \to \mathbb N\cup \{\infty\}$ be the maximal number $n$ for which $\hat{\mathcal G}^{n+1}_{\eta}(v,w)$ is well defined, that is,
\begin{equation}\label{eq:taudef}
    \tau(v,w) := \inf\{n \in \mathbb N: \hat{\mathcal G}_{\eta}^n(v,w) \notin V^*\}.\footnote{Of course, $\inf \emptyset := \infty$.}
\end{equation}
One can easily check that for $(v,w) \in V^*$ and $1 \leq n \leq \tau(v,w)$, we have
\begin{equation}\label{eq:GDlift}
    \hat{\mathcal G}^n_{\eta}(v,w) = \chi^{-1}(\mathcal G^n_{\eta}(\chi(v,w))).
\end{equation}
The corresponding statement for SGD will be given in Section \ref{sec:RDS}  once the appropriate notation has been introduced (cf.~\eqref{eq:SGDlift}).

Since all points in $\mathcal M$ are fixed points for $\hat{\mathcal G}_{\eta}, \hat{\mathcal G}_{\eta,1}, \dots \hat{\mathcal G}_{\eta, N}$ and by Lemma \ref{lemm:localco} (ii), we have 
$\hat V \cup (\mathbb R^{D-N} \times \{0\}) \subseteq V^*$
and
\begin{align*}
    \hat{\mathcal G}_{\eta}(v,0) &= (v,0) &&\text{and} &\hat{\mathcal G}_{\eta,i}(v,0) &= (v,0).
\end{align*}
Furthermore, using the chain rule, equations \eqref{eq:conjGD} and \eqref{eq:conjSGD} as well as Lemma \ref{lemm:localco} (iii), we can compute 
\begin{align}\label{eq:GhatJac}
    \hat{\mathcal G}'_{\eta}(0,0) &=  \mathds 1_{D}- \frac{\eta}{N} \begin{pmatrix}
       0&0\\0&G_{x^*}
    \end{pmatrix} &&\text{and} &\hat{\mathcal G}'_{\eta,i}(0,0) &= \mathds 1_{D}- \eta \begin{pmatrix}
       0&0\\0&G_{x^*,[i]}
    \end{pmatrix},
\end{align}
where the right-hand sides are both block matrices with dimensions $((D-N)+N)\times ((D-N)+N)$\revision{\footnote{\revision{i.e.~the rows and the columns have been partitioned into two blocks of size $(D-N)$ and size $N$.}}}.
\begin{lemma}\label{lemm:approx}
    For every $\delta > 0$, there exists an open neighborhood $V_\delta \subseteq V^*$, such that for each $(v,w) \in V_\delta$ and each $i \in [N]$, we have
    \begin{equation}\label{ineq:GDapprox}
        \left\| \hat{\mathcal G}_{\eta}(v,w) - \left[\hat{\mathcal G}'_{\eta}(0,0)\right]\begin{pmatrix}v\\w\end{pmatrix}\right\| \leq \delta\|w\|,
    \end{equation}
    as well as
    \begin{equation}\label{ineq:SGDapprox}
        \left\| \hat{\mathcal G}_{\eta,i}(v,w) - \left[\hat{\mathcal G}'_{\eta,i}(0,0)\right]\begin{pmatrix}v\\w\end{pmatrix}\right\| \leq \delta\|w\|.
    \end{equation}
\end{lemma}
Note that this lemma would hold trivially if the right-hand sides of \eqref{ineq:GDapprox} and \eqref{ineq:SGDapprox} were replaced by the term $\delta \|(v,w)\|$.
\begin{proof}
    It is sufficient to consider each of the functions $\hat{\mathcal G}_{\eta},  \hat{\mathcal G}_{\eta,1}, \dots,  \hat{\mathcal G}_{\eta,N}$ separately, find a neighborhood $V_\delta$ in which the corresponding inequality \eqref{ineq:GDapprox}, respectively \eqref{ineq:SGDapprox}, holds and conclude the proof by choosing the intersection over all these $V_\delta$. In the following, we will only consider $\hat{\mathcal G}_{\eta}$ and will show how to find an $V_\delta$ such that \eqref{ineq:GDapprox} holds, as the proofs for $\hat{\mathcal G}_{\eta,1}, \dots,  \hat{\mathcal G}_{\eta,N}$ work analogously.

    Let $0 \in V_\delta \subseteq V^*$ be an open, convex neighborhood, such that
    $$\left\|\partial_w \hat{\mathcal G}_{\eta}(v,w) - \partial_w \hat{\mathcal G}_{\eta}(0,0)\right\| \leq \delta, ~ \forall\, (v,w) \in V_\delta.$$
    Here $\partial_w \hat{\mathcal G}_{\eta}(v,w) \in \mathbb R^{D \times N}$ denotes the partial Jacobian with respect to the latter $N$ components.
    We have
    \begin{align*}
         \hat{\mathcal G}_{\eta}(v,w) - \left[\hat{\mathcal G}'_{\eta}(0,0)\right]\begin{pmatrix}v\\w\end{pmatrix}&= \hat{\mathcal G}_{\eta}(v,0)+\int_0^1 \left[\partial_w \hat{\mathcal G}_{\eta}(v,tw)\right]w \,\dd t- (v,0) -\left[\partial_w \hat{\mathcal G}_{\eta}(0,0)\right]w\\
         &= \int_0^1 \left[\partial_w \hat{\mathcal G}_{\eta}(v,tw) - \partial_w \hat{\mathcal G}_{\eta}(0,0)\right]w \,\dd t
    \end{align*}
    and for $(v,w) \in V_\delta$ in particular
    $$\left\|\hat{\mathcal G}_{\eta}(v,w) - \left[\hat{\mathcal G}'_{\eta}(0,0)\right]\begin{pmatrix}v\\w\end{pmatrix}\right\| \leq \int_0^1 \left\|\partial_w \hat{\mathcal G}_{\eta}(v,tw) - \partial_w \hat{\mathcal G}_{\eta}(0,0)\right\|\|w\| \,\dd t \leq \delta \|w\|.$$
    This finishes the proof.
\end{proof}

\subsection{Gradient Descent - the Stable Case}\label{sec:gdstable}
In this section we will prove Theorem \ref{theo:mainGD} (i).
\begingroup
\def\thetheorem{\ref{theo:mainGD} (i)}
\begin{theorem}
Let $x^* \in \mathcal M$ with $\mu(x^*) < 0$. Then $x^* \in \operatorname{supp}\left(X_{\lim}^{\operatorname{GD}}\right)$. 
\end{theorem}
\addtocounter{theorem}{-1}
\endgroup

In the following, we will let $\Pi_v \in \mathbb R^{(D-N)\times D}$ and $\Pi_w \in \mathbb R^{(D-N)\times N}$ be the matrices that project a vector $(v,w)^t \in \mathbb R^D$ onto its $v$-, respectively $w$-component, i.e.~in block-matrix form
\begin{align*}
    \Pi_v &:= \begin{pmatrix}\mathds 1_{D-N}& 0\end{pmatrix}&&\text{and}& \Pi_w &:= \begin{pmatrix}0&\mathds 1_{N}\end{pmatrix}.
\end{align*}
\begin{proof}[of Theorem \ref{theo:mainGD} (i)]
    Suppose $\mu(x^*)< 0$ and let $U \subseteq \mathcal M$ be some neighborhood of $x^*$. Our goal is to prove $\mathbb P(X^{\operatorname{GD}}_{\lim} \in U) > 0$.   
    Choose $\delta>0$ such that $e^{\mu(x^*)}+\delta =: \gamma <1$, let $V_\delta$ be the corresponding neighborhood given by Lemma \ref{lemm:approx}. Let $R_\delta > 0$ be some radius, such that
    $$\mathcal B_{\mathbb R^{D-N}}(R_\delta) \times \mathcal B_{\mathbb R^{N}}(R_\delta) \subseteq V_\delta.$$
    We assume without loss of generality that $U$ has the form
    \begin{equation}\label{eq:deltaBallinV}
    U = \chi\left(\mathcal B_{\mathbb R^{D-N}}(R) \times \{0\}\right)
    \end{equation}
    for some $0<R<R_\delta$.
    Similarly to the definition of $\tau$, we define a map $\tau_\delta: V_\delta \to \mathbb N\cup \{\infty\}$ by
    \begin{equation}\label{eq:tauDelta}
        \tau_\delta(v,w) := \inf\left\{n \in \mathbb N: \hat{\mathcal G}_{\eta}^n(v,w) \notin V_\delta\right\}.
    \end{equation}
    Note that $\tau_\delta(v,w)\leq \tau(v,w)$. 

    As a consequence of Lemma \ref{lemm:approx} and the expression \eqref{eq:NTKmuDef} for $\mu(x^*)$, for all $(v,w) \in V_\delta$, we have
    \begin{align}
        \left\|\Pi_w \hat{\mathcal G}_{\eta}(v,w)\right\| &\leq \left\|\left[\Pi_w\hat{\mathcal G}'_{\eta}(0,0)\right]\begin{pmatrix}v\\w\end{pmatrix}\right\|+\left\|\Pi_w\left(\hat{\mathcal G}_{\eta}(v,w) - \left[\hat{\mathcal G}'_{\eta}(0,0)\right]\begin{pmatrix}v\\w\end{pmatrix}\right)\right\|\nonumber\\
        &\leq\left\|\left(\mathds 1_N - \frac{\eta}{N} G_{x^*}\right)w\right\|+\left\|\hat{\mathcal G}_{\eta}(v,w) - \left[\hat{\mathcal G}'_{\eta}(0,0)\right]\begin{pmatrix}v\\w\end{pmatrix}\right\|\nonumber\\
        &\leq e^{\mu(x^*)} \|w\| + \delta \|w\| = \gamma \|w\|.\label{ineq:gdOpNorm}
    \end{align}
    Using this bound inductively, for all $(v,w) \in V_\delta$ and $1 \leq n \leq \tau_\delta(v,w)$, we get
    \begin{equation}\label{ineq:wbound}
        \left\|\Pi_w \hat{\mathcal G}^n_{\eta}(v,w)\right\| \leq \gamma^n\|w\|.
    \end{equation}
   Recalling equation~\eqref{eq:GhatJac} and using Lemma \ref{lemm:approx}  implies
    $$\left\|\Pi_v\hat{\mathcal G}_{\eta}(v,w) - v\right\| = \left\|\Pi_v\left(\hat{\mathcal G}_{\eta}(v,w) - \left[\hat{\mathcal G}'_{\eta}(0,0)\right]\begin{pmatrix}v\\w\end{pmatrix}\right)\right\|\leq \delta \|w\|,$$
    for all $(v,w) \in V_\delta$ and thus also for all $1 \leq n < \tau_\delta(v,w)$,
    \begin{equation}\label{ineq:cauchy}
        \left\|\Pi_v\hat{\mathcal G}^{n+1}_{\eta}(v,w) - \Pi_v\hat{\mathcal G}^n_{\eta}(v,w)\right\| \leq \delta\left\|\Pi_w\hat{\mathcal G}^n_{\eta}(v,w)\right\| \leq \delta \gamma^n \|w\|.
    \end{equation}
    With this, we can bound
    \begin{align}
        \left\|\Pi_v\hat{\mathcal G}^n_{\eta}(v,w)\right\|&\leq\|v\|+\sum_{m = 1}^n \left\|\Pi_v\hat{\mathcal G}^m_{\eta}(v,w) - \Pi_v\hat{\mathcal G}^{m-1}_{\eta}(v,w)\right\|\nonumber\\
        &\leq \|v\|+\sum_{m = 1}^n \delta \gamma^{m-1} \|w\| \leq \|v\|+\frac{\delta}{1-\gamma}\|w\|,\label{ineq:vbound}
    \end{align}
    for all $(v,w) \in V_\delta$ and $1 \leq n+1 < \tau_\delta(v,w)$. Now set 
    \begin{align*}
        R_v &:= \frac{R}{2}&&\text{and} &R_w &= \min \left(\frac{(1-\gamma)R}{2\delta}, R_\delta\right).
    \end{align*}
    \begin{figure}
    \centering
    \begin{overpic}[scale=.25,tics=10,, width = 0.5\textwidth]{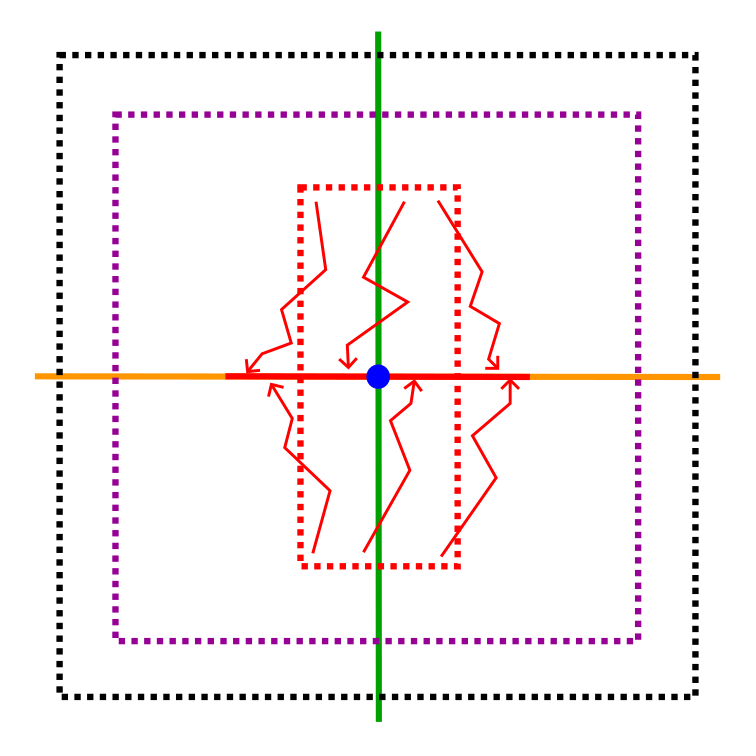}
        \put(51,52){\color{blue}$(0,0)$}
        \put(60,95){$\mathcal B_{\mathbb R^{D-N}}(\hat R) \times \mathcal B_{\mathbb R^N}(\hat R)$}
        \put(76,17){\color{Plum}$V_\delta$}
        \put(62,70){\color{red} $\mathcal B_{\mathbb R^{D-N}}( R_v) \times \mathcal B_{\mathbb R^N}(R_w)$}
        \put(0,43){\color{red}$\mathcal B_{\mathbb R^{D-N}}(R) \times \{0\}$}
        \put(75,52){\color{orange}$\mathbb R^{D-N}\times \{0\}$}
        \put(51,19){\color{Green}$\{0\}\times \mathbb R^N$}
    \end{overpic}
    \caption{Schematic representation of the proof of Theorem \ref{theo:mainGD} (i). The red arrows show possible sample trajectories.}
    \label{fig:gdstable}
\end{figure}
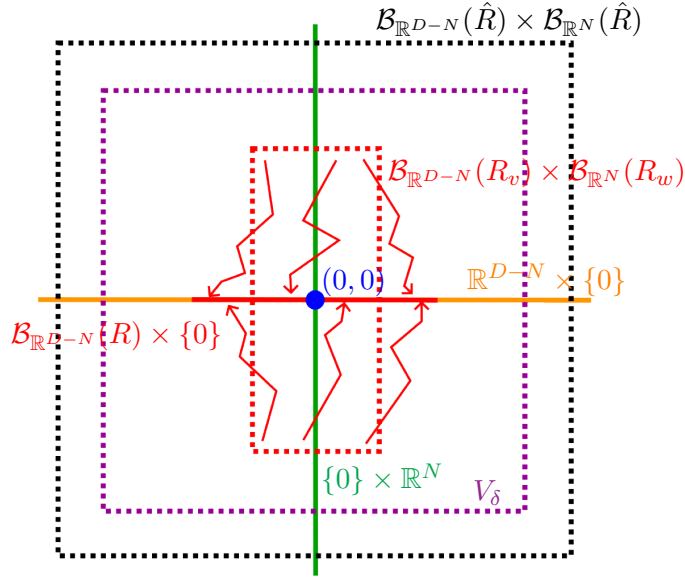
    Suppose for some $(v,w) \in \mathcal B_{\mathbb R^{D-N}}(R_v) \times \mathcal B_{\mathbb R^N}(R_w)$, we have $\tau_\delta(v,w) < \infty$. Then by definition of $\tau_\delta$, we have $\hat{\mathcal G}^{\tau_\delta(v,w)}_{\eta}(v,w)\notin V_\delta$, so by \eqref{eq:deltaBallinV} in particular 
    \begin{align}\label{eq:notinUdelta}
        \left\|\Pi_v \hat{\mathcal G}^{\tau_\delta(v,w)}_{\eta}(v,w)\right\| &\geq R_\delta &&\text{or}& \left\|\Pi_w \hat{\mathcal G}^{\tau_\delta(v,w)}_{\eta}(v,w)\right\| &\geq R_\delta.
    \end{align}
    However, \eqref{ineq:vbound} implies
    $$ \left\|\Pi_v \hat{\mathcal G}^{\tau_\delta(v,w)}_{\eta}(v,w)\right\| \leq \|v\|+\frac{\delta}{1-\gamma}\|w\| <R_v+\frac{\delta}{1-\gamma}R_w \leq R \leq R_\delta$$
    and \eqref{ineq:wbound} implies
    $$\left\|\Pi_w \hat{\mathcal G}^{\tau_\delta(v,w)}_{\eta}(v,w)\right\| \leq \gamma^{\tau_\delta(v,w)} \|w\| \leq \|w\| < R_w \leq R_\delta,$$
    contradicting \eqref{eq:notinUdelta}. Thus $\tau_\delta(v,w) = \infty$, for all $(v,w) \in \mathcal B_{\mathbb R^{D-N}}(R_v) \times \mathcal B_{\mathbb R^N}(R_w)$. Now \eqref{ineq:cauchy} implies that $\left(\Pi_v\hat{\mathcal G}^{n}_{\eta}(v,w)\right)$ is a Cauchy sequence and \eqref{ineq:vbound} shows that
    $$\left\|\lim_{n \to \infty}\Pi_v\hat{\mathcal G}^n_{\eta}(v,w)\right\| \leq \|v\|+\frac{\delta}{1-\gamma}\|w\| <R_v+\frac{\delta}{1-\gamma}R_w \leq R.$$
    Furthermore, \eqref{ineq:wbound} shows $\Pi_w\hat{\mathcal G}^n_{\eta}(v,w) \to 0$. Thus, for each $(v,w) \in \mathcal B_{\mathbb R^{D-N}}(R_v) \times \mathcal B_{\mathbb R^N}(R_w)$ the sequence $\left(\hat{\mathcal G}^{n}_{\eta}(v,w)\right)$ converges with
    $$\lim_{n \to \infty} \hat{\mathcal G}^{n}_{\eta}(v,w) \in \mathcal B_{\mathbb R^{D-N}}(R) \times \{0\}.$$
    Let $\tilde U = \chi\left(\mathcal B_{\mathbb R^{D-N}}(R_v) \times \mathcal B_{\mathbb R^N}(R_w)\right)$. Suppose $X_0^{\operatorname{GD}} \in \tilde U$. By \eqref{eq:GDlift} and continuity of $\chi$ we have
    \begin{align*}
        X^{\operatorname{GD}}_{\lim} &= \lim_{n \to \infty}\mathcal G^{n}_{\eta}(X_0^{\operatorname{GD}}) = \lim_{n \to \infty}\chi\left(\hat{\mathcal G}^{n}_{\eta}\left(\chi^{-1}(X_0^{\operatorname{GD}})\right)\right)\\
        &= \chi\left(\lim_{n \to \infty}\hat{\mathcal G}^{n}_{\eta}\left(\chi^{-1}(X_0^{\operatorname{GD}})\right)\right) \in \chi(\mathcal B_{\mathbb R^{D-N}}(R) \times \{0\}) = U,
    \end{align*}
    so $X_0^{\operatorname{GD}} \in \tilde U$ implies $X^{\operatorname{GD}}_{\lim} \in U$.
    Since $\tilde U$ is an open set, we have
    $$\mathbb P(X^{\operatorname{GD}}_{\lim} \in U) \geq \mathbb P\left(X_0^{\operatorname{GD}}\in \tilde U\right) > 0,$$
    completing the proof.
\end{proof}

\subsection{Gradient Descent - the Unstable Case}\label{sec:gdcenterstable}
In this section, we will prove Theorem \ref{theo:mainGD} (ii).
\begingroup
\def\thetheorem{\ref{theo:mainGD} (ii)}
\begin{theorem}
Let $x^* \in \mathcal M$ with $\mu(x^*) > 0$. Then $x^* \notin \operatorname{supp}\left(X_{\lim}^{\operatorname{GD}}\right)$. 
\end{theorem}
\addtocounter{theorem}{-1}
\endgroup

In the case $\mu(x^*) < 0$ considered in the previous section, all the eigenvalues of {$\mathds 1_N - \eta G_{x^*}$} lie strictly within the unit circle. As a consequence, we were able to construct an open neighborhood of $x^*$, such that for every initial condition in that neighborhood, GD converges to some $X_{\lim}^{\operatorname{GD}} \in \mathcal M$ near $x^*$. In the case $\mu(x^*)>0$ we consider in this section, we only know that some eigenvalue of {$\mathds 1_N - \eta G_{x^*}$} lies strictly outside the unit circle while there could still be eigenvalues on or strictly inside the unit circle. As a consequence, one should not expect to be able to construct a neighborhood $U\subseteq \mathbb R^D$ of $x^*$, such that GD does not converge to some $X_{\lim}^{\operatorname{GD}} \in \mathcal M$ near $x^*$ for any initial condition in $U$. Instead, we will construct a neighborhood $U\subseteq \mathbb R^D$ of $x^*$, such that GD does not converge to some $X_{\lim}^{\operatorname{GD}} \in \mathcal M$ near $x^*$ for Lebesgue-almost any initial condition in $U$. In fact, we will show that the set of initial conditions in $U$ for which GD does converge to some $X_{\lim}^{\operatorname{GD}} \in \mathcal M$ near $x^*$ is contained in a lower dimensional manifold called the \emph{center-stable manifold}\footnote{While the set, we call center-stable manifold is indeed a $C^1$-manifold (cf.~Theorem 6.2.8 in \citealp{KatokHasselblatt95}), we will only show that it is the graph of a Lipschitz continuous function.}. While for the preceding and all subsequent sections it is convenient to consider vectors in $\mathbb R^D$ as consisting of a $D-N$-dimensional tangential part $v$ and an $N$-dimensional transversal part $w$, here it will be more convenient to consider vectors in $\mathbb R^D$ as consisting of a center-stable part $v_-$ corresponding to the eigenvalues strictly inside or on the unit circle and an unstable part $v_+$ corresponding to the eigenvalues strictly outside the unit circle.

For the remainder of this section, consider a fixed $x^* \in \mathcal M$ with $\mu(x^*) > 0$. Let $\bar \mu_1, \dots, \bar \mu_D$ be the eigenvalues of be the eigenvalues of $\hat{\mathcal G}'(0,0)$ appearing according to their multiplicity\footnote{Since $\hat{\mathcal G}'_\eta(0,0)$ is symmetric, the algebraic and geometric multiplicities coincide.} and ordered by their absolute values, i.e.
$$|\bar \mu_1| \geq \dots \geq |\bar \mu_D|.$$
Note that the eigenvalue 1 must appear at least with multiplicity $D-N$, corresponding to the tangential part and $\log|\bar \mu_1| = \mu(x^*) > 0$, which implies $|\bar \mu_1|>1$. Let $D_+>1$ be the number of eigenvalues with absolute value greater than 1 and let $D_- = D-D_+$ be the number of eigenvalues with absolute value less or equal to 1, counted with multiplicity in both cases. Furthermore, let $\bar S \in \mathbb R^{D \times D}$ be an invertible matrix, such that
$$\hat{\mathcal G}'_\eta(0,0) = \bar S \begin{pmatrix}\bar \mu_1&&\\&\ddots&\\&&\bar \mu_D\end{pmatrix} \bar S^{-1} = \bar S \begin{pmatrix}A_+&0\\0&A_-\end{pmatrix} \bar S^{-1},$$
where $A_+ \in \mathbb R^{D_+ \times D_+}$ and $A_- \in \mathbb R^{D_-\times D_-}$ are given by 
\begin{align*}
    A_+ &:= \begin{pmatrix}\bar \mu_1&&\\&\ddots&\\&&\bar \mu_{D_+}\end{pmatrix}&&\text{and}&A_- &:= \begin{pmatrix}\bar \mu_{D_++1}&&\\&\ddots&\\&&\bar \mu_{D}\end{pmatrix}.
\end{align*}
Note that $\|A_-\| = 1$ and that $A_+$ is invertible with $\|A_+^{-1}\|^{-1} = |\bar\mu_{D_+}|>1$, where $\bar \mu_{D_+}$ is the eigenvalue with the smallest absolute value that is still larger than 1.
Finally, let $\bar {\mathcal G}_\eta: \mathbb R^{D_+}\times \mathbb R^{D_-} \to \mathbb R^{D_+}\times \mathbb R^{D_-}$ given by
$$\bar {\mathcal G}_\eta(v_+, v_-) = \bar S \hat{\mathcal G}_\eta \left(\bar S^{-1}\begin{pmatrix}v_+\\v_-\end{pmatrix}\right).$$
The following is a version of the center-stable manifold theorem. 
\begin{theorem}[Center-Stable Manifold]\label{theo:centerstable}
    There exist open neighborhoods $0 \in V_-\subseteq \mathbb R^{D_-}$, $0 \in V_+\subseteq \mathbb R^{D_+}$ and a map $\beta^*: V_- \to V_+$ such that for any $(v_+, v_-) \in (V_+ \times V_-) \setminus \operatorname{graph}(\beta^*)$, there exists some $n \in \mathbb N,$ such that $\bar {\mathcal G}_\eta^n(v_+, v_-) \notin V_+ \times V_-$.
\end{theorem}

If $\bar {\mathcal G}_\eta$ is a local diffeomorphism at 0, this result is standard (see e.g.~ chapter 6.2 in \citealp{KatokHasselblatt95}). However, in our setting, it is possible that $\bar \mu_i = 0$ for some $i \in [D]$. Still, Theorem \ref{theo:centerstable} follows from the arguments in \cite{KatokHasselblatt95} with mild modifications. Alternatively, stable/unstable/center manifold theorems in non-locally diffeomorphic setting have also been obtained by reducing the problem to the locally diffeomorphic case by an abstract method called inverse limits (see, e.g.,~\citealp{RuelleShub80} or \citealp{Qianetal09}). 

In order to prove Theorem \ref{theo:mainGD} (ii), we require the following lemma. Since the equivalent statement will also be useful to prove the instability of SGD, we formulate it to cover both GD and SGD.
\begin{lemma}\label{lemm:nullsets}
    Let $A \subseteq \mathbb R^D$ be a Lebesgue-null set. Then 
    $$\mathbb P\left( \exists n \in \mathbb N_0, \text{ s.t. } X_n^{\operatorname{GD/SGD}} \in A\right) = 0.$$
\end{lemma}
\begin{proof}
    By Hypothesis \ref{assu:init}, we have $\mathbb P(X_0^{\operatorname{GD/SGD}} \in A) = \nu(A) = 0$. By Hypothesis \ref{assu:nonsing}, we also have $\mathbb P(X_n^{\operatorname{GD/SGD}}\in A) = 0$, for every $n \geq 1$ and thus 
    $$\mathbb P\left( \exists n \in \mathbb N_0, \text{ s.t. } X_n^{\operatorname{GD/SGD}} \in A\right) = 0.$$
\end{proof}

\begin{proof}[of Theorem \ref{theo:mainGD} (ii)]
    Suppose $\mu(x^*) > 0$. Let $x^* \in U \subseteq \mathbb R^D$ be the open neighborhood given by 
    $$U := \chi(\bar S (V_+ \times V_-)),$$
    where $V_+$ and $V_-$ are the neighborhoods given by Theorem \ref{theo:centerstable}. Suppose $\omega \in \Omega$ is such that $X_{\lim}^{\operatorname{GD}}(\omega) \in U \cap \mathcal M$. Then there must exist an $m \in \mathbb N$, such that $\mathcal G_\eta^n (X_{m}^{\operatorname{GD}}(\omega)) = X_{m+n}^{\operatorname{GD}(\omega)} \in U$ for all $n \in \mathbb N_0$. Equivalently, we must have $\bar {\mathcal G}_\eta^n (\bar S^{-1} \chi^{-1}(X_{m}^{\operatorname{GD}}(\omega))) \in V_+ \times V_-$, for all $n \in \mathbb N_0$. By Theorem \ref{theo:centerstable} this means that we can only have $X_{\lim}^{\operatorname{GD}}(\omega) \in U \cap \mathcal M$, if there exists an $m \in \mathbb N$ such that $X_m^{\operatorname{GD}}(\omega) \in \chi(\bar S \operatorname{graph}(\beta^*))$. However, since $\chi$ is a diffeomorphism and $\bar S$ an invertible matrix, the set $\chi(\bar S \operatorname{graph}(\beta^*))$ has Lebesgue measure zero. By Lemma \ref{lemm:nullsets}, this implies
    $$\mathbb P(X_{\lim}^{\operatorname{GD}}(\omega) \in U \cap \mathcal M) = 0,$$
    completing the proof.
\end{proof}

\subsection{Random Dynamical System Framework for SGD}\label{sec:RDS}
While in sections \ref{sec:gdstable} and \ref{sec:gdcenterstable} we studied the dynamics of gradient descent, the rest of the paper will be concerned with establishing analogous results for stochastic gradient descent. For this, we will formulate SGD as a \emph{random dynamical system} (see \citealp{Bible} for a general introduction to the theory). Formally, for some appropriate probability space $(\Omega, \mathcal F, \mathbb P)$, we will introduce a map
\begin{align*}
    \varphi: \mathbb N_0 \times \Omega \times \mathbb R^D &\to \mathbb R^D,\\
    (n, \omega, x) &\mapsto \varphi^{(n)}_\omega(x),
\end{align*}
where evaluation of the map $\varphi^{(n)}_\omega$ should correspond to applying $n$ iterations of SGD with $\omega$ serving as the seed for the random choices of training examples. 
We expect the map $\varphi$ to satisfy
\begin{equation}\label{eq:cocycle}
     \varphi^{(n+m)}_\omega(x) = \varphi^{(m)}_{\theta^n \omega}\left(\varphi^{(n)}_\omega(x)\right),
\end{equation}
where $\theta^n \omega$ is the seed for the same training examples, but shifted by $n$ steps.
Put more clearly, if $\omega$ generates the sequence of training examples $(\xi_1, \xi_2, \dots)\in [N]^{\mathbb N}$, then $\theta^n\omega$ should generate the sequence $(\xi_{n+1}, \xi_{n+2}, \dots)\in [N]^{\mathbb N}$. Since the training examples are chosen independently with identical distributions, the sequences $(\xi_1, \xi_2, \dots)$ and $(\xi_{n+1}, \xi_{n+2}, \dots)$ are equal in distribution. In other words, the map $\theta: \Omega \to \Omega$ leaves the probability measure $\mathbb P$ invariant, i.e.
\begin{equation}\label{eq:metricds}
    \mathbb P(E) = \mathbb P\left(\theta^{-1}(E)\right), ~\forall\, E \in \mathcal F.
\end{equation}
A map $\theta: \Omega \to \Omega$ satisfying \eqref{eq:metricds} is called \emph{metric dynamical system}. Equation \eqref{eq:cocycle} is known as the \emph{cocycle property} and a map $\varphi$ satisfying it is called a \emph{cocycle} over $\theta$. A pair $(\theta, \varphi)$, consisting of a metric dynamical system and a cocycle over it, is called a random dynamical system (cf.~Defintion 1.1.1 in \citealp{Bible}).

Since $\Omega$ should encode the randomness both at initialization and during training, we let 
\begin{align*}
    \Omega &:= \mathbb R^D \times [N]^{\mathbb N} &&\text{and}& \mathcal F &:= \mathfrak B(\mathbb R^D) \otimes \mathfrak P([N])^{\otimes \mathbb N},
\end{align*}
where $\mathfrak P([N])$ denotes the power set of $[N]$. If $\beta$ denotes the uniform measure on $[N]$ we can define the probability measure $\mathbb P: \mathcal F \to [0,1]$ by
$$\mathbb P := \nu \otimes \beta^{\otimes \mathbb N}.$$
Thus, the elements $\omega \in \Omega$ have the form $\omega = (\omega_{\operatorname{init}}, \omega_1, \omega_2, \dots)$ and the random variables $X^{\operatorname{SGD}}_0, \xi_1, \xi_2, \dots$ used in SGD (cf.~Section \ref{subs:sgd}) can be formally defined by
\begin{align*}
X^{\operatorname{SGD}}_0(\omega)  &:= \omega_{\operatorname{init}} &&\text{and}& \xi_n(\omega) := \omega_n.
\end{align*}
The shift operator $\theta: \Omega \to \Omega$ can now be defined by
$$\theta: \omega = (\omega_{\operatorname{init}}, \omega_1, \omega_2, \dots) \mapsto \theta \omega := (\omega_{\operatorname{init}}, \omega_2, \omega_3, \dots).$$
Clearly $\theta$ satisfies \eqref{eq:metricds} and thus defines a metric dynamical system on $\Omega$. Furthermore $\theta$ is ergodic with respect to the sub-sigma algebra $\hat{\mathcal F} := \sigma(\xi_1, \xi_2, \dots)$, i.e.
$$\theta^{-1}(E) = E \Rightarrow \mathbb P(E) \in \{0,1\}, ~\forall\, E \in \hat{\mathcal F}.$$
Recall (cf.~\eqref{eq:SGD} and \eqref{eq:Getai}) that the random variables $X^{\operatorname{SGD}}_1, X^{\operatorname{SGD}}_2, \dots$ can be defined recursively by
$$X^{\operatorname{SGD}}_{n+1} (\omega) := \mathcal G_{\eta, \xi_{n+1}(\omega)}(X^{\operatorname{SGD}}_n(\omega)), ~\forall\, n \geq 0.$$
The map $\varphi: \mathbb N_0 \times \Omega \times \mathbb R^D \to \mathbb R^D$ will be defined by
\begin{align}\label{eq:phiAsCon}
    \varphi^{(0)}_\omega(x) &:= x &&\text{and}& \varphi^{(n)}_\omega(x) &:= \left[\mathcal G_{\eta,\xi_n(\omega)}\circ \dots \circ \mathcal G_{\eta,\xi_1(\omega)}\right](x).
\end{align}
It is easy to check that $\varphi$ satisfies \eqref{eq:cocycle}. Thus $\varphi$ is a cocylce over $\theta$ and $(\theta, \varphi)$ a random dynamical system. The random variables $X_n^{\operatorname{SGD}}$ can now be expressed as
$$X_n^{\operatorname{SGD}}(\omega) = \varphi_\omega^{(n)}\left(X_0^{\operatorname{SGD}}(\omega)\right).$$
Note that the points $x \in \mathcal M$ are fixed points of $\varphi$, that is,
\begin{equation}\label{eq:fixedpoints}
    \varphi_\omega^{(n)}(x) = x,~\forall \, n \in \mathbb N_0, \omega \in \Omega, x \in \mathcal M.
\end{equation}

In order to introduce an RDS framework for the linearization of SGD (cf.~Section \ref{sec:linstab}), for SGD in local coordinates (cf.~Section \ref{sec:localco}) and for linearized SGD in local coordinates, we will again fix a global minimum $x^* \in \mathcal M$ for the remainder of this section. We define a map $\Phi: \mathbb N_0 \times \Omega \to \mathbb R^{D\times D}$ by
\begin{equation}\label{def:Phi}
    \Phi: (n, \omega) \mapsto \Phi_\omega^{(n)}:= \mathrm D_x \varphi_\omega^{(n)}(x^*),
\end{equation}
where $\mathrm D_x \varphi_\omega^{(n)}(x^*)$ denotes the Jacobian of the map $\varphi_\omega^{(n)}$. Applying the chain rule to \eqref{eq:phiAsCon} and using \eqref{eq:fixedpoints}, we get
\begin{align*}
    \Phi_\omega^{(0)} &= \mathds 1_D &&\text{and}& \Phi_\omega^{(n)} &= \mathcal G'_{\eta,\xi_n(\omega)}(x^*) \dots \mathcal G'_{\eta,\xi_1(\omega)}(x^*).
\end{align*}
From this, it is easy to see that $\Phi$ is a \emph{matrix cocycle}, i.e.~that it satisfies
$$\Phi^{(n+m)}_\omega = \Phi^{(m)}_{\theta^n\omega}\Phi^{(n)}_\omega.$$
Since the matrices $\mathcal G'_{\eta, 1}, \dots, \mathcal G'_{\eta, N}$ all respect the splitting $\mathbb R^D = \mathcal T(x^*) \oplus \mathcal N(x^*)$ with $\mathcal G'_{\eta, i}|_{\mathcal T(x^*)} = \operatorname{Id}_{\mathcal T(x^*)}$ (cf.~Section \ref{sec:linstab}), the same holds true for $\Phi_\omega^{(n)}$ for all $\omega \in \Omega$ and all $n \in \mathbb N_0$.

Recall that $\chi: \hat V \to \hat U$, introduced in Lemma \ref{lemm:localco}, defines a diffeomorphism from a neighborhood $\hat V$ of the origin to a neighborhood $\hat U$ of $x^*$. Furthermore, we introduced a neighborhood $V^* \subseteq \hat V$ (cf.~\eqref{eq:ustar}), which allowed us to locally lift the maps $\mathcal G_{\eta, 1}, \dots, \mathcal G_{\eta, n}$ via $\chi$ to the maps $\hat{\mathcal G}_{\eta, 1}, \dots, \hat{\mathcal G}_{\eta, n}: V^* \to \hat V$ defined by
$$\hat{\mathcal G}_{\eta,i}(v,w) := \chi^{-1}(\mathcal G_{\eta,i}(\chi(v,w))).$$
Analogously to the definition of $\tau: V^* \to \mathbb N \cup \{\infty\}$ in Section \ref{sec:localco} (cf.~\eqref{eq:taudef}), we introduce a map $\tau: \Omega \times V^* \to \mathbb N \cup \{\infty\}$ by
$$\tau_\omega(v,w) := \inf\left\{n \in \mathbb N: \left[\hat{\mathcal G}_{\eta,\xi_n(\omega)}\circ \dots \circ\hat{\mathcal G}_{\eta,\xi_1(\omega)}\right](v,w) \notin V^*\right\}.$$
This allows us to define a map $\psi: \mathbb N_0 \times \Omega \times V^* \supseteq D_\psi \to \hat V$, where 
$$D_\psi:= \left\{(n, \omega, v,w) \in \mathbb N_0 \times \Omega \times V^*: n \leq \tau_\omega(v,w)\right\},$$
by
$$\psi_\omega^{(n)}(v,w) := \left[\hat{\mathcal G}_{\eta,\xi_n(\omega)}\circ \dots \circ\hat{\mathcal G}_{\eta,\xi_1(\omega)}\right](v,w).$$
Clearly, $\psi$ satisfies the \emph{local cocycle property}, meaning 
$$\psi^{(n+m)}_\omega(x) = \psi^{(m)}_{\theta^n \omega}\left(\psi^{(n)}_\omega(x)\right)$$
whenever the left-hand side is well defined, i.e.~whenever $(n+m, \omega, v, w) \in D_\psi$. This turns the pair $(\theta, \psi)$ into a \emph{local random dynamical system} (cf.~Definition 1.2.1 in \citealp{Bible}). The local cocycle $\psi$ can be seen as the local lift of $\varphi$ via $\chi$, as one can easily check that
\begin{equation}\label{eq:SGDlift}
    \psi^{(n)}_\omega(v,w) = \chi^{-1}\left(\varphi^{(n)}_\omega(\chi(v,w))\right),
\end{equation}
whenever $n \leq \tau_\omega(v,w)$.

Similarly to the definition of $\Phi$, we introduce a matrix cocycle $\Psi: \mathbb N_0 \times \Omega \to \mathbb R^{N \times N}$ by 
\begin{align}\label{eq:PsiDef}
    \Psi_\omega^{(0)} &:= \mathds 1_N &&\text{and}& \Psi_\omega^{(n)}&= \left(\mathds 1_N - \eta G_{x^*,[\xi_n]}\right) \dots \left(\mathds 1_N - \eta G_{x^*,[\xi_1]}\right).
\end{align}
Note that, as opposed to $\psi$, the matrix cocycle $\Psi$ is defined globally. This is possible since $\tau_\omega(0,0) = \infty$, for all $\omega \in \Omega$. Also by \eqref{eq:GhatJac}, we can express the Jacobian of $\psi_\omega^{(n)}$ at the origin by
$$\mathrm D_{(v,w)} \psi_\omega^{(n)}(0,0) = \begin{pmatrix}\mathds 1_{D-N}&0 \\0& \Psi_\omega^{(n)}\end{pmatrix}.$$
Alternatively, this can also be seen by differentiating \eqref{eq:PsiDef} and using Lemma \ref{lemm:localco}, as well as \eqref{eq:conjSGD}. Furthermore, differentiating \eqref{eq:SGDlift} at (0,0) yields the identity
\begin{equation}\label{eq:PhiViaPsi}
    \Phi_\omega^{(n)} = \begin{pmatrix}A&S_{x^*}\end{pmatrix}\begin{pmatrix}\mathds 1_{D-N}&0 \\0& \Psi_\omega^{(n)}\end{pmatrix}\begin{pmatrix}A&S_{x^*}\end{pmatrix}^{-1},
\end{equation}
where $\begin{pmatrix}A&S_{x^*}\end{pmatrix} \in \mathbb R^{D \times D}$ is the matrix from Lemma \ref{lemm:localco} (iii).
The following corollary is a simple reformulation of Lemma \ref{lemm:approx}.
\begin{corollary}\label{coro:approx}
     For every $\delta > 0$, there exists a neighborhood $(0,0) \in V_\delta \subseteq V^*$, such that for each $(v,w) \in V_\delta$, we have
    \begin{equation*}
        \left\| \psi^{(1)}_\omega(v,w) - \left(v, \Psi^{(1)}_\omega w\right)\right\| \leq \delta\|w\|.
    \end{equation*}
\end{corollary}
Finally, we can reformulate \eqref{eq:NTKlambdaDef} in terms of $\Psi$ to get the expression
\begin{equation}\label{eq:lambdaPsi}
    \lambda(x^*) = \inf_{n \in \mathbb N} \frac{1}{n} \mathbb E \left[\log \left\|\Psi_\omega^{(n)}\right\|\right]
\end{equation}
for the Lyapunov exponent. By Kingman's sub-additive ergodic theorem\footnote{Note that the fact that $\lambda(x^*)$ does not depend on $\omega$ requires $\theta$ to be ergodic, which it is not. However, the cocycle $\Psi$ is measurable with respect to the sub-sigma algebra $\hat{\mathcal F} = \sigma(\xi_1, \xi_2, \dots)$. Thus we may consider the probability space $(\Omega, \hat{\mathcal F}, \mathbb P|_{\hat{\mathcal F}})$, in which $\theta$ is ergodic, when applying the sub-additive ergodic theorem.}, we have
\begin{equation}\label{eq:lambdaLim}
    \lambda(x^*) = \lim_{n \to \infty} \frac{1}{n} \log \left\|\Psi_\omega^{(n)}\right\|,
\end{equation}
for almost every $\omega \in \Omega$.

\subsection{Stochastic Gradient Descent - the Stable Case}\label{sec:sgdstable}
In this section we will prove Theorem \ref{theo:mainSGD} (i).
\begingroup
\def\thetheorem{\ref{theo:mainSGD} (i)}
\begin{theorem}
 Let $x^* \in \mathcal M$ with $\lambda(x^*) < 0$ and suppose
    \begin{equation}\label{etacond2}
    \|\nabla_x\mathfrak F(x^*,y_i)\|^2 \neq \frac{1}{\eta},
\end{equation}
for every $i \in [N]$. Then $x^* \in \operatorname{supp}\left(X_{\lim}^{\operatorname{SGD}}\right)$. 
\end{theorem}
\addtocounter{theorem}{-1}
\endgroup

Since this is the SGD equivalent of Theorem \ref{theo:mainGD} (i), the proof will be similar. However, there is a major obstacle. A crucial ingredient in the proofs of Theorem \ref{theo:mainGD} (i) is the expression $\|\mathds 1_N - \eta G_{x^*}\| = e^{\mu(x^*)}$ (cf.~\eqref{ineq:gdOpNorm}), which allowed us to derive the bound (cf.~\eqref{ineq:wbound})
\begin{equation}\label{ineq:wboundCopy}
    \left\|\Pi_w \hat{\mathcal G}^n_{\eta}(v,w)\right\| \leq \gamma^n\|w\|.
\end{equation}
For stochastic gradient descent, we only have a bound of the form 
$$\left\|\Psi_\omega^{(n)}\right\| \leq \tilde C_\delta(\omega) e^{n(\lambda(x^*) + \delta)},$$
where for any $\delta > 0$, $\tilde C_\delta$ is a random variable which is finite almost surely. This can be seen as a consequence of $\eqref{eq:lambdaLim}$. Random variables similar to $\tilde C_\delta$ are sometimes called \emph{Oseledec regularity functions} in the literature. If $\tilde C_\delta(\omega)$ was almost surely bounded by some deterministic constant, it would be possible to derive the equivalent statement to \eqref{ineq:wboundCopy} with an additional factor on the right-hand side.
Unfortunately, the Oseledec regularity function $\tilde C_\delta$ will generally be unbounded. This is one of the defining features of so-called \emph{non-uniform hyperbolicity}. Instead, we will derive the equivalent statement to \eqref{ineq:wboundCopy} by essentially using the upper semi-continuity of the Lyapunov exponent (cf.~Lemma \ref{lemm:oseReg} below). This allows for a proof of Theorem \ref{theo:mainSGD} (i) along the lines of the proof of Theorem \ref{theo:mainGD} (i). It should be noted that this does not save the proof of Theorem $\ref{theo:mainSGD}$ (ii). Thus an entirely different approach will be presented in sections \ref{sec:semiGr}-\ref{sec:sgdunstable}.
\begin{lemma}\label{lemm:oseReg}
    Let $x^* \in \mathcal M$ be a point for which \eqref{etacond2} is satisfied. For each $\gamma> e^{\lambda(x^*)}$, there exists a $\delta >0$ and a random variable $C_\gamma: \Omega \to (0, \infty]$ such that $\mathbb P(C_\gamma(\omega) < \infty) = 1$ and 
    \begin{equation*}
        \left\|\Pi_w \psi_\omega^{(n)}(v,w)\right\| \leq C_\gamma(\omega)\gamma^n \|w\|, ~ \forall\, \omega \in \Omega,\,(v,w) \in V_\delta,\, n \leq \tau_{\delta, \omega}(v,w),
    \end{equation*}
    where $V_\delta$ is the neighborhood from Lemma \ref{lemm:approx}/Corollary \ref{coro:approx} and $\tau_{\delta,\omega}: V_\delta \to \mathbb N \cup \{\infty\}$ the map given by 
    $$\tau_{\delta, \omega}(v,w) := \inf\left\{n \in \mathbb N: \psi^{(n)}_\omega(v,w) \notin V_\delta\right\}.$$
\end{lemma}

\begin{proof}
    Let $x^* \in \mathcal M$ be such that \eqref{etacond2} holds and $\gamma> e^{\lambda(x^*)}$. Choose an $\varepsilon>0$ such that $e^{\lambda(x^*)+2\varepsilon} \leq \gamma$. By $\eqref{eq:lambdaPsi}$, there exists an $n^* \in \mathbb N$, such that
    \begin{equation}\label{ineq:logToNStar}
        \mathbb E \left[\log \left\|\Psi^{(n^*)}_\omega\right\|\right] < n^* (\lambda(x^*)+\varepsilon).
    \end{equation}
    Since \eqref{etacond2} holds, the matrix $\Psi^{(1)}_\omega$ is invertible for every $\omega \in \Omega$ and we may define constants $0<K_1<K_2<\infty$ by
    \begin{align}\label{eq:K1def}
        K_1 &:= \inf_{\omega \in \Omega} \left\|\left(\Psi^{(1)}_\omega\right)^{-1}\right\|^{-1} = \min_{i \in [N]} \left\|\left(\mathds 1_N - \eta G_{x^*,[i]}\right)^{-1}\right\|^{-1},\\
        K_2 &:= \sup_{\omega \in \Omega} \left\|\Psi^{(1)}_\omega\right\| = \max_{i \in [N]} \left\|\mathds 1_N - \eta G_{x^*,[i]}\right\|.\label{eq:K2def} 
    \end{align}
    By the cocycle property we have
    \begin{equation}\label{ineq:Kbound}
        K_1^n \|w\|\leq \left\|\Psi^{(n)}_\omega w\right\| \leq K_2^n \|w\|.
    \end{equation}
 For some $\delta>0$ to be determined later, let $V_\delta$ and $\tau_{\delta, \omega}$ be as described in the formulation of the Lemma. By Corollary \ref{coro:approx}, we have the na\"ive bound
 \begin{align*}
     \left\|\Pi_w \psi^{(1)}_\omega (v,w)\right\| \leq \left\|  \Pi_w\psi^{(1)}_\omega(v,w) - \Psi^{(1)}_\omega w\right\| + \left\|\Psi^{(1)}_\omega w\right\|\leq (K_1+\delta) \|w\|
 \end{align*}
 for all $\omega \in \Omega$ and $(v,w) \in V_\delta$. Thus we also have
 \begin{equation}\label{ineq:naivebound}
     \left\|\Pi_w \psi^{(n)}_\omega (v,w)\right\| \leq (K_2+\delta)^n \|w\|,
 \end{equation}
for all $\omega \in \Omega$, $(v,w) \in V_\delta$ and $n \leq \tau_{\delta, \omega}(v,w)$.
 
 Suppose that $\omega \in \Omega$ and $(v, w) \in V_\delta$ satisfy $\tau_{\delta, \omega}(v,w) \geq n^*$. By the cocycle properties for $\psi$ and $\Psi$, the inequality \eqref{ineq:naivebound}, Corollary \ref{coro:approx} and the inequality \eqref{ineq:Kbound}, we have
 \begin{align*}
     &\phantom{\leq}\left\|\Pi_w\psi^{(n^*)}_\omega(v,w) - \Psi^{(n^*)}_\omega w\right\| \\
     &\leq \sum_{n = 1}^{n^*}\left\|\Pi_w\psi^{(n^*-n-1)}_{\theta^{n+1}\omega}\left(\psi^{(1)}_{\theta^{n}\omega}\left(v, \Psi^{(n)}_\omega w\right)\right) -\Pi_w\psi^{(n^*-n-1)}_{\theta^{n+1}\omega}\left(v, \Psi^{(1)}_{\theta^{n}\omega}\Psi^{(n)}_\omega w\right)\right\|\\
     &\leq \sum_{n = 1}^{n^*}(K_2+\delta)^{n^*-n-1}\left\|\Pi_w\psi^{(1)}_{\theta^{n}\omega}\left(v, \Psi^{(n)}_\omega w\right) - \Psi^{(1)}_{\theta^{n}\omega}\Psi^{(n)}_\omega w\right\|\\
     &\leq \sum_{n = 1}^{n^*}\delta (K_2+\delta)^{n^*-n-1}\left\|\Psi^{(n)}_\omega w\right\|\\
     &\leq \sum_{n = 1}^{n^*}\delta (K_2+\delta)^{n^*-n-1}K_1^{n^*-n}\left\|\Psi^{(n^*)}_\omega w\right\|.
 \end{align*}
We now fix $\delta>0$ to have a small enough value, s.t.
$$\sum_{n = 1}^{n^*}\delta (K_2+\delta)^{n^*-n-1}K_1^{n^*-n} \leq \varepsilon n^*.$$
Thus we have
$$\left\|\Pi_w\psi^{(n^*)}_\omega(v,w) - \Psi^{(n^*)}_\omega w\right\| \leq \varepsilon n^*\left\|\Psi^{(n^*)}_\omega w\right\|,$$
for all  $\omega \in \Omega$ and $(v, w) \in V_\delta$ with $\tau_{\delta, \omega}(v,w) \geq n^*$. In that case we also have
\begin{align}
    \left\|\Pi_w \psi^{(n^*)}_\omega (v,w)\right\| &\leq \left\|\Psi_\omega^{(n^*)}w\right\| + \left\|\Pi_w\psi^{(n^*)}_\omega(v,w) - \Psi^{(n^*)}_\omega w\right\|\nonumber\\
    &\leq (1+\varepsilon n^*)\left\|\Psi_\omega^{(n^*)}w\right\|\leq \hat C(\omega) \|w\|,\label{ineq:refinedBound}
\end{align}
where we define $\hat C(\omega) := (1+\varepsilon n^*)\left\|\Psi_\omega^{(n^*)}\right\|$. By \eqref{ineq:logToNStar}, we have
\begin{align}
    \mathbb E\left[\log\left(\hat C(\omega)\right)\right] &= \mathbb \log(1+\varepsilon n^*)+\mathbb E\left[\log\left(\Psi_\omega^{(n^*)}\right)\right]\nonumber\\
    &\leq\varepsilon n^* + n^*(\lambda(x^*)+\varepsilon) = n^*(\lambda(x^*)+2\varepsilon)<n^*\log(\gamma).\label{ineq:logMom}
\end{align}
Next we introduce another random variable $\tilde C: \Omega \to \mathbb R \cup \{\infty\}$ by
$$\tilde C(\omega) := \sup_{\ell \in \mathbb N} \sum_{k=0}^{\ell-1} \left[ \log\left(\hat C(\theta^{kn^*}\omega)\right)- n^*\log(\gamma)\right].$$
Note that the random variable $\hat C$ is measurable with respect to the sigma algebra $\mathcal F_{n^*}:= \sigma(\xi_1, \dots, \xi_{n^*})$ and that the random variables $\left(\hat C(\theta^{kn^*}\omega)\right)_{k \in \mathbb N_0}$ are thus independent and by $\theta$-invariance (cf.~\eqref{eq:metricds}) identically distributed. By the strong law of large numbers, we have, for almost every $\omega \in \Omega$,
$$\lim_{\ell \to \infty} \frac{1}{\ell}\sum_{k=0}^{\ell-1} \left[ \log\left(\hat C(\theta^{kn^*}\omega)\right)- n^*\log(\gamma)\right] = \mathbb E\left[\log\left(\hat C(\omega)\right)\right] - n^*\log(\gamma)<0$$
and thus in particular $\mathbb P(\tilde C(\omega) < \infty) = 1$. By definition, the random variable $\tilde C$ can be used to obtain the bound
\begin{equation}\label{ineq:Ctilde}
    \sum_{k=0}^{\ell-1}\log\left(\hat C(\theta^{kn^*}\omega)\right)  \leq \tilde C(\omega) + \ell n^*\log(\gamma).
\end{equation}
Combining \eqref{ineq:refinedBound} and \eqref{ineq:Ctilde}), we get
\begin{align}
    \left\|\Pi_w \psi^{(\ell n^*)}_\omega (v,w)\right\| &\leq \left[\prod_{k = 0}^{\ell -1} \hat C\left(\theta^{kn^*}\omega\right)\right]\|w\|\nonumber\\
    &= \exp\left(\sum_{k = 0}^{\ell -1} \log\left(\hat C\left(\theta^{kn^*}\omega\right)\right)\right) \|w\|\nonumber\\
    &\leq e^{\tilde C(\omega) + \ell n^*\log(\gamma)} \|w\|= e^{\tilde C(\omega)}\gamma^{\ell n^*} \|w\|,\label{ineq:nstarstep}
\end{align}
for all $\omega \in \Omega$, $(v,w) \in V_\delta$ and $\ell \in \mathbb N$ with $\tau_{\delta, \omega}(v,w) \geq \ell n^*$. 

Now let $\omega \in \Omega$, $(v,w) \in V_\delta$ and $n \in \mathbb N$ with $\tau_{\delta, \omega}(v,w) \geq n$. Let $\ell \in \mathbb N_0$ and $0 \leq k \leq n^*-1$ be such that $n = \ell n^*+k$. Using the cocycle property and the bounds \eqref{ineq:naivebound} and \eqref{ineq:nstarstep}, we obtain
\begin{align*}
    \left\|\Pi_w \psi^{(n)}_\omega (v,w)\right\| &= \left\|\Pi_w \psi^{(k)}_{\theta^{\ell n^*}\omega}\left(\psi^{(\ell n^*)}_\omega (v,w)\right)\right\|\\
    &\leq (K_2+\delta)^k \left\|\Pi_w \psi^{(\ell n^*)}_\omega (v,w)\right\|\\
    &\leq (K_2+\delta)^ke^{\tilde C(\omega)}\gamma^{\ell n^*} \|w\|\\
    &= \left(\frac{K_2+\delta}{\gamma}\right)^k e^{\tilde C(\omega)}\gamma^{\ell n^* + k}\|w\|\\
    &\leq \left(\frac{K_2+\delta}{\gamma}\right)^{n^*-1}e^{\tilde C(\omega)} \gamma^{n}\|w\|.
\end{align*}
Thus we can define the random variable $C_\gamma: \Omega \to \mathbb R \cup \infty$ by
$$C_\gamma(\omega):= \left(\frac{K_2+\delta}{\gamma}\right)^{n^*-1}e^{\tilde C(\omega)},$$
satisfying $\mathbb P(C_\gamma(\omega) < \infty) = 1$.
\end{proof}
With Lemma \ref{lemm:oseReg} in place, we can prove Theorem \ref{theo:mainSGD} (i) analogously to the proof of Theorem \ref{theo:mainGD} (i) presented in Section \ref{sec:gdstable}.
\begin{proof}[of Theorem \ref{theo:mainSGD} (i)]
    Suppose $\lambda(x^*)< 0$ and let $U \subseteq \mathcal M$ be some neighborhood of $x^*$. Our goal is to prove $\mathbb P(X^{\operatorname{SGD}}_{\lim} \in U) > 0$. Choose $\gamma \in \mathbb R$, such that $e^{\lambda(x^*)}< \gamma <1$ and let $\delta>0$ and $C_\gamma: \Omega \to \mathbb R \cup \infty$ be such that the conclusion of Lemma \ref{lemm:oseReg} holds, i.e.~such that $\mathbb P(\mathbb C_\gamma(\omega) < \infty) = 1$ and that we have
    \begin{equation}\label{ineq:wbound2}
        \left\|\Pi_w \psi_\omega^{(n)}(v,w)\right\| \leq C_\gamma(\omega)\gamma^n \|w\|,
    \end{equation}
    for all $\omega \in \Omega$, $(v,w) \in V_\delta$ and $n \leq \tau_{\delta, \omega}(v,w)$. 
    This bound will serve as the equivalent to \eqref{ineq:wbound} in the proof of Theorem \ref{theo:mainGD} (i).
    Let $R_\delta >0$ be some radius, such that
    $$\mathcal B_{\mathbb R^{D-N}}(R_\delta) \times \mathcal B_{\mathbb R^N} (R_\delta) \subseteq V_\delta.$$
    We may assume without loss of generality that $U$ has the form
    $$U = \chi\left(\mathcal B_{\mathbb R^{D-N}}(R) \times \{0\}\right)$$
    for some $R\leq R_\delta$.
    As a consequence of Corollary \ref{coro:approx} we get
    $$\left\|\Pi_v\psi_\omega^{(1)}(v,w) - v\right\| \leq \delta \|w\|,$$
    for all $(v,w) \in V_\delta$ and thus also for all $1 \leq n < \tau_{\delta, \omega}(v,w)$,
    \begin{equation}\label{ineq:cauchy2}
        \left\|\Pi_v\psi_\omega^{(n+1)}(v,w) - \Pi_v\psi_\omega^{(n)}(v,w)\right\| \leq \delta\left\|\Pi_w\psi_\omega^{(n)}(v,w)\right\| \leq \delta C_\gamma(\omega) \gamma^n \|w\|.
    \end{equation}
    With this, we can bound
    \begin{align}
        \left\|\Pi_v\psi_\omega^{(n)}(v,w)\right\|&\leq\|v\|+\sum_{m = 0}^{n-1} \left\|\Pi_v\psi_\omega^{(m+1)}(v,w) - \Pi_v\psi_\omega^{(m)}(v,w)\right\|\nonumber\\
        &\leq \|v\|+\sum_{m = 0}^{n-1} \delta C_\gamma(\omega) \gamma^m \|w\| \nonumber\\
        &\leq \|v\|+\frac{\delta C_\gamma(\omega)}{1-\gamma}\|w\|,\label{ineq:vbound2}
    \end{align}
    for all $(v,w) \in V_\delta$ and $1 \leq n+1 < \tau_{\delta, \omega}(v,w)$. Now set 
    \begin{align*}
        R_v &:= \frac{R}{2}&&\text{and} &R_w(\omega) &= \min \left(\frac{(1-\gamma)R}{2\delta C_\gamma(\omega)}, \frac{R_\delta}{C_\gamma(\omega)}\right).
    \end{align*}
    Suppose for some $\omega \in \Omega$ with $C_\gamma(\omega) < \infty$ and some $(v,w) \in \mathcal B_{\mathbb R^{D-N}}(R_v) \times \mathcal B_{\mathbb R^N}(R_w(\omega))$, we have $\tau_{\delta, \omega}(v,w) < \infty$. Then by definition $\psi_\omega^{(\tau_{\delta, \omega}(v,w))}(v,w)\notin V_\delta$, so in particular
    \begin{align}\label{eq:notinUdelta2}
        \left\|\Pi_v \psi_\omega^{(\tau_{\delta, \omega}(v,w))}(v,w)\right\| &\geq R_\delta &&\text{or}& \left\|\Pi_w \psi_\omega^{(\tau_{\delta, \omega}(v,w))}(v,w)\right\| &\geq R_\delta.
    \end{align}
    However, \eqref{ineq:vbound2} implies
    $$ \left\|\Pi_v \psi_\omega^{(\tau_{\delta, \omega}(v,w))}(v,w)\right\| \leq \|v\|+\frac{\delta C_\gamma(\omega)}{1-\gamma}\|w\| <R_v+\frac{\delta C_\gamma(\omega)}{1-\gamma}R_w(\omega) \leq R \leq R_\delta$$
    and \eqref{ineq:wbound2} implies
    $$\left\|\Pi_w \psi_\omega^{(\tau_{\delta, \omega}(v,w))}(v,w)\right\| \leq C_\gamma(\omega)\gamma^{\tau_{\delta, \omega}(v,w)} \|w\| \leq C_\gamma(\omega)\|w\| < C_\gamma R_w \leq R_\delta,$$
    contradicting \eqref{eq:notinUdelta2}. Thus $\tau_{\delta, \omega}(v,w) = \infty$, for all $(v,w) \in \mathcal B_{\mathbb R^{D-N}}(R_v(\omega)) \times \mathcal B_{\mathbb R^N}(R_w(\omega))$. Now \eqref{ineq:cauchy2} implies that $\left(\psi_\omega^{(n)}(v,w)\right)$ is a Cauchy sequence and \eqref{ineq:vbound2} shows that
    $$\left\|\lim_{n \to \infty}\Pi_v\psi_\omega^{(n)}(v,w)\right\| \leq \|v\|+\frac{\delta C_\gamma(\omega)}{1-\gamma}\|w\| <R_v+\frac{\delta C_\gamma(\omega)}{1-\gamma}R_w(\omega) \leq R.$$
    Furthermore, \eqref{ineq:wbound2} shows $\Pi_w\psi_\omega^{(n)}(v,w) \to 0$. Thus, for each $(v,w) \in \mathcal B_{\mathbb R^{D-N}}(R_v) \times \mathcal B_{\mathbb R^N}(R_w(\omega))$ the sequence $\left(\psi_\omega^{(n)}(v,w)\right)$ converges with
    $$\lim_{n \to \infty} \psi_\omega^{(n)}(v,w) \in \mathcal B_{\mathbb R^{D-N}}(R) \times \{0\}.$$
    Let $\tilde U(\omega) = \chi\left(\mathcal B_{\mathbb R^{D-N}}(R_v) \times \mathcal B_{\mathbb R^N}(R_w(\omega))\right)$ whenever $C_\gamma(\omega) < \infty$. Suppose $X_0^{\operatorname{SGD}}(\omega) \in \tilde U(\omega)$. By \eqref{eq:SGDlift} and continuity of $\chi$ we have
    \begin{align*}
        X^{\operatorname{SGD}}_{\lim}(\omega) &= \lim_{n \to \infty}\varphi_\omega^{(n)}(X^{\operatorname{SGD}}_0(\omega)) = \lim_{n \to \infty}\chi\left(\psi_\omega^{(n)}\left(\chi^{-1}(X^{\operatorname{SGD}}_0(\omega))\right)\right)\\
        &= \chi\left(\lim_{n \to \infty}\psi_\omega^{(n)}\left(\chi^{-1}(X_0^{\operatorname{SGD}}(\omega)\right)\right) \in \chi(\mathcal B_{\mathbb R^{D-N}}(R) \times \{0\}) = U,
    \end{align*}
    so $X_0^{\operatorname{SGD}}(\omega) \in \tilde U(\omega)$ implies $X^{\operatorname{SGD}}_{\lim}(\omega) \in U$.
    By construction $\tilde U$ is measurable with respect to $\sigma(\xi_1, \xi_2, \dots)$ and thus independent of $X_0^{\operatorname{SGD}}$. Therefore we have
    $$\mathbb P(X^{\operatorname{SGD}}_{\lim}(\omega) \in U) \geq \mathbb P \left(X_0^{\operatorname{SGD}}(\omega) \in \tilde U(\omega)\right) = \mathbb E\left[\nu\left(\tilde U(\omega)\right)\right].$$
    Since $\tilde U$ is a non-empty open set almost surely, by Hypothesis \ref{assu:init} we get
    $$\mathbb P(X^{\operatorname{SGD}}_{\lim}(\omega) \in U) \geq \mathbb E\left[\nu\left(\tilde U(\omega)\right)\right] > 0,$$
    completing the proof.

\end{proof}

\subsection{Generated Matrix semigroups}\label{sec:semiGr}
It remains to prove Theorem \ref{theo:mainSGD} (ii). The proof of Theorem \ref{theo:mainSGD} (i) in the previous section was of a \emph{quenched} nature. One might expect that the best approach to proving Theorem \ref{theo:mainSGD} (ii) is to construct $\omega$-wise center-stable manifolds similar to the ones constructed in Section \ref{sec:gdcenterstable}. While an invariant manifold theory (Pesin theory) has been developed for random dynamical systems (see e.g.~\citealp{LiuQian1995} or \citealp{Bible}), these results only provide center-stable manifolds for single points $x \in \mathcal M$. For the argument in Section \ref{sec:gdcenterstable} it was crucial to have a center-stable manifold for an open subset of $\mathcal M$. The authors are not aware of any method to construct such a random center-stable manifold.

In the following, we will present an \emph{annealed} argument. Instead of showing that, given $\lambda(x^*) > 0$, for almost every $\omega$ the points which converge to any $X_{\lim} \in \mathcal M$ near $x^*$ form a $\nu$
-null set, we will show that $\nu$-almost every initial condition $x_0 \in \mathbb R^D$ does not converge to any $X_{\lim} \in \mathcal M$ near $x^*$ almost surely. Both statements are equivalent to Theorem \ref{theo:mainSGD} (ii) by Fubini's theorem, but they are different in flavor. The former is a statement on the  deterministic dynamics for a fixed $\omega$, while the latter concerns the stochastic behavior of a Markov process. The advantage of the stochastic approach is that under the conditions we impose, namely regularity of $x^*$ in the sense of Definition \ref{def:regular}, the center-stable manifolds get ``washed away" by the randomness: while for every $\omega$ there might exist a manifold of initial conditions which still converge to $\mathcal M$ near $x^*$, for every initial condition $X_0 \notin \mathcal M$, the probability of converging to some $X_{\lim} \in \mathcal M$ near $x^*$ is zero. 

We will show this by constructing a \emph{Lyapunov function} (see e.g.~\citealp{Tobias}) defined on a neighborhood of $x^*$ which goes to infinity near $\mathcal M$. This is inspired by previous work on the instability of invariant subspaces for stochastic differential equations
(cf.~\citealp{Baxendale1991, BaxendaleStroock, BedrossianBlumenthalPunshon2022, BlumenthalCotiGvalani2023, CotiZelatiHairer}).
These works rely on conditions of Hörmander type to establish the existence of a spectral gap in the so-called projective process. Such a spectral gap is excluded by the discrete nature of our problem. Instead, we use an argument due to \cite{lepage} (cf.~also \citealp{RMP} for a survey in English) to find a spectral gap in a different topology. Le Page's argument needs the matrix semigroup on which the linear cocycle $\Psi$ is supported to satisfy two algebraic properties, namely being contracting and strongly irreducible (cf.~Definition \ref{def:csi} below). In this section, we will show that they follow from the regularity of $x^*$.

For some $x^* \in \mathcal M$, we denote the support of the matrix-valued random variable $\Psi^{(n)}_\bullet$ by $\mathcal S_n(x^*)$, i.e.
$$\mathcal S_n(x^*):= \operatorname{supp}\left(\Psi_\bullet^{(n)}\right) \subset \mathbb R^{N\times N}.$$
From the definition of $\Psi$ \eqref{eq:PsiDef}, one can readily see that
$$\mathcal S_n(x^*) = \left\{\left(\mathds 1_N -\eta G_{x^*,[\xi_n]}\right)\dots \left(\mathds 1_N -\eta G_{x^*,[\xi_1]}\right) : (\xi_1, \dots, \xi_n) \in [N]^n\right\}.$$
Furthermore, we denote the total support of $\Psi$ by $\mathcal S(x^*)$, i.e. 
$$\mathcal S(x^*) := \bigcup_{n = 0}^\infty \mathcal S_n(x^*) \subset \mathbb R^{N\times N}.$$
Clearly, $\mathcal S(x^*)$ is a matrix semigroup with unity, generated by $\mathcal S_1(x^*)$, i.e.~$\mathcal S(x^*)$ contains exactly those matrices, which can be expressed as the product of an arbitrary number of elements in $\mathcal S_1(x^*)$, including the empty product, which is defined to be the identity matrix.

Recall from Definition \ref{def:regular}, that a point $x^* \in \mathcal M$ is called regular, if for every $i \in \mathbb N$, we have
        \begin{equation*}
            [G_{x^*}]_{i,i} = \|\nabla_x\mathfrak F(x^*,y_i)\|^2 \notin \left\{\frac{1}{\eta}, \frac{2}{\eta}\right\}, 
        \end{equation*}
        and if there exists no proper subset $\emptyset \subsetneq \mathcal A \subsetneq [N]$, such that 
        $$[G_{x^*}]_{i,j} = \nabla_x\mathfrak F(x^*,y_i) \cdot \nabla_x\mathfrak F(x^*,y_j) = 0,~\forall \, i \in \mathcal A,\, j \in [N] \setminus \mathcal A.$$
\begin{proposition}
    Let $x^*\in \mathcal M$ be regular. Then $\mathcal S(x^*) \subseteq \gl(N, \mathbb R)$.
\end{proposition}
\begin{proof}
    Since $x^*\in \mathcal M$ is regular, in particular $$\|\nabla_x\mathfrak F(x^*,y_i)\|^2 \neq\frac{1}{\eta},~\forall\, i \in [N].$$
    Thus $\mathcal S_1(x^*) \subseteq \gl(N, \mathbb R)$ and since $\mathcal S(x^*)$ is generated by $\mathcal S_1(x^*)$ also $\mathcal S(x^*) \subseteq \gl(N, \mathbb R)$.
\end{proof}
\begin{definition}\label{def:csi}
A matrix semigroup of invertible matrices $\mathcal S \subseteq \gl(N, \mathbb R)$ is called,
\begin{enumerate}
\item[(i)] \emph{contracting}, if there exists a sequence $(M_n)_{n \in \mathbb N} \subset \mathcal S$, such that $$\lim_{n \to \infty}\frac{M_n}{\|M_n\|} = M,$$ for some rank-1 matrix $M \in \mathbb R^{N \times N}$.
    \item[(ii)]\emph{strongly irreducible}, if for every proper linear subspace $\{0\} \subsetneq W \subsetneq \mathbb R^N$ the set of subspaces $\{MW: M \in \mathcal S\}$ contains infinitely many elements.
\end{enumerate}
\end{definition}
\begin{lemma}\label{lemm:CSI}
    Let $x^* \in \mathcal M$ be a regular point with $\lambda(x^*) > 0$. Then $\mathcal S(x^*) \subseteq \gl(N, \mathbb R)$ is both contracting and strongly irreducible.
\end{lemma}
\begin{proof}
    Let $x^* \in \mathcal M$ be a regular point with $\lambda(x^*) > 0$. For ease of notation, we will write $G := G_{x^*}$ and $G_{[i]} := G_{x^*,[i]}$. We start by proving that $\mathcal S(x^*)$ is contracting.
    From the original definition \eqref{eq:lambdaDef} of $\lambda(x^*)$, we get the inequality
    \begin{align*}
        0 < \lambda(x^*) \leq \mathbb E\left[\log\left\|\mathcal G_{\eta, \xi_1}'(x^*)\right\|\right] &\leq \max_{i \in [N]}\log\left\|\mathcal G_{\eta, i}'(x^*)\right\| \\
        &= \max_{i \in [N]}\log\left\|\left(\mathds 1_D - \eta\nabla_x \mathfrak F(x^*, y_{i})\nabla_x \mathfrak F(x^*, y_{i})^t \right)|_{\mathcal N(x^*)}\right\|.
    \end{align*}
    In particular, there must exist an $i^* \in [N]$ with 
    $$\left|1 - \eta \left\|\nabla_x \mathfrak F(x^*, y_{i^*})\right\|^2\right| = \left|1 - \eta G_{i^*,i^*}\right| > 1.$$
    Without loss of generality, we assume $i^* = 1$. Consider the sequence $(M_n) \in \mathcal S(x^*)^{\mathbb N}$ given by
    $$M_n := \left(\mathds 1_N -\eta G_{[1]}\right)^n.$$
    Recall that $G_{[1]}$ is the matrix $G$ with all but the first row replaced by zeros. Thus $[G_{[1]}]$ is a rank 1 matrix with non-trivial eigenvalue $G_{1,1}$ and the eigenvalues of $\mathds 1_N -\eta G_{[1]}$ are $\mu_1 = 1 - \eta G_{1,1}$ with multiplicity 1 and $\mu_2 = 1$ with multiplicity $N-1$. 
    Using basic finite-dimensional spectral theory, $\mathds 1_N -\eta G_{[1]}$ can be decomposed as $\mathds 1_N -\eta G_{[1]} = A + B$, where $A$ is a rank-1 matrix with $A^n = \mu_1^{n-1} A$, $AB = BA = 0$ and 
    \begin{equation}\label{eq:fracAB}
        \lim_{n \to \infty} \frac{\|B^n\|}{\|A^n\|} = 0.
    \end{equation}
    Now we can compute
    $$M_n = \left(\mathds 1_N -\eta G_{[1]}\right)^n = (A+B)^n = A^n + B^n$$
    and thus 
    \begin{align*}
        \lim_{n \to \infty }\frac{M_n}{\|M_n\|} &= \lim_{n \to \infty} \frac{A^n + B^n}{\left\|A^n + B^n\right\|} = \lim_{n \to \infty} \frac{\left\|A^n\right\|}{\left\|A^n+B^n\right\|}\frac{A^n}{\left\|A^n\right\|} +  \frac{B^n}{\left\|A^n + B^n\right\|}.
    \end{align*}
    As a consequence of \eqref{eq:fracAB}, we have 
    \begin{align*}
        \lim_{n \to \infty} \frac{\left\|A_n\right\|}{\left\|A^n+B^n\right\|} &=1 && \text{and}& \lim_{n \to \infty} \frac{B^n}{\left\|A^n + B^n\right\|} &= 0
    \end{align*}
    and hence 
    $$\lim_{n \to \infty }\frac{M_n}{\|M_n\|} = \lim_{n \to \infty } \frac{A^n}{\left\|A^n\right\|} = \lim_{n \to \infty } \frac{\mu_1^{n-1}A}{\mu_1^{n-1}\left\|A\right\|} = \frac{A}{\|A\|},$$
    which is indeed a rank-1 matrix.

    In order prove that $\mathcal S(x^*)$ is strongly irreducible, consider a proper linear subspace $\{0\} \subsetneq W \subsetneq \mathbb R^N$ for wich we intend to show that $\{MW: M \in \mathcal S\}$ contains infinitely many elements.
    Consider the sets $\mathcal A, \mathcal B \subseteq [N]$ given by 
    \begin{align*}
        \mathcal A &:= \{i \in [N] : e_i \in W\}&&\text{and}&\mathcal B &:= \{i \in [N] : G e_i \in W^\bot\},
    \end{align*}
    where $e_i$ denotes the $i$-th unit vector. Since $G$ is positive definite\footnote{Recall that $G$ is the Gram matrix of the neural tangent kernel.}, we have $e_i^t Ge_i > 0$ for all $i$ and thus $\mathcal A \cap \mathcal B = \emptyset$. Also, we have $G_{i,j} = e_i^t G e_j = 0$, for each $i\in \mathcal A$ and $j \in \mathcal B$. By the assumption that $x^*$ is regular, this implies $\mathcal B \neq [N] \setminus \mathcal A$ and thus $\mathcal A \cup \mathcal B \neq [N]$. In other words, there must exists an $i^* \in [N]$, such that $e_{i^*} \notin W$ and $G e_{i^*} \notin W^\bot$. The latter implies that there must exists some $w^* \in W$ such that $e_i^tGw^* =w^{*t}Ge_i \neq 0$. Again, we assume without loss of generality that $i^* = 1$. Consider the sequence of subspaces $(W_n) \in \{MW: M \in \mathcal S\}^{\mathbb N_0}$ given by $W_n := (\mathds 1_N - \eta G_{[1]})^n W$ and let
    $$\kappa_n := \sup_{w \in W_n\setminus \{0\}} \frac{\|G_{[1]}w\|}{\left\|w- \frac{1}{G_{1,1}}G_{[1]}w\right\|}.$$
    Note that since the term in the supremum only depends on the direction of $w$ and not on $\|w\|$, it is sufficient to take the supremum over the unit ball and by compactness the supremum must be attained. Since $G_{[1]}w$ is always a multiple of $e_1$ and $e_1 \notin W_0 = W$, the denominator is always non-zero for $n = 0$. Furthermore, $e_1^tGw^* \neq 0$ implies $G_{[1]}w^* \neq 0$ and we have $0 < \kappa_0 < \infty$. Also, using $G_{[1]}^2 = G_{i,i} G_{[1]}$, we get
    \begin{align*}
    G_{[1]}(\mathds 1_N - \eta G_{[1]})w &= (1-\eta G_{1,1}) G_{[1]}w \ \text{ and}\\
    (\mathds 1_N - \eta G_{[1]}) w- \frac{1}{G_{1,1}}G_{[1]}(\mathds 1_N - \eta G_{[1]})w &= w- \frac{1}{G_{1,1}}G_{[1]}w.
    \end{align*}
    Applying this iteratively, allows us to compute
    \begin{align*}
        \kappa_n &= \sup_{w \in W\setminus \{0\}} \frac{\|G_{[1]}(\mathds 1_N - \eta G_{[1]})^nw\|}{\left\|(\mathds 1_N - \eta G_{[1]})^n w- \frac{1}{G_{1,1}}G_{[1]}(\mathds 1_N - \eta G_{[1]})^nw\right\|}\\
        &=\sup_{w \in W\setminus \{0\}} |1-\eta G_{1,1}|^n\frac{\|G_{[1]}w\|}{\left\|w- \frac{1}{G_{1,1}}G_{[1]}w\right\|}\\
        &= |1-\eta G_{1,1}|^n \kappa_0.
    \end{align*}
    Since $G_{1,1} \neq \frac{2}{\eta}$, we have $|1-\eta G_{1,1}| \neq 1$ and the sequence $(\kappa_n)$ consists of pairwise distinct elements. Thus, in particular, the sequence $(W_n)$ consists of pairwise distinct subspaces, completing the proof.
\end{proof}

\subsection{Construction of Lyapunov Functions}\label{sec:LyapFun}
The goal of this section is, given \revision{a global minimum $x^*$ with} $\lambda(x^*) > 0$, to construct a Lyapunov function $F^*: \mathbb R^N\setminus\{0\} \to [0, \infty)$ which goes to infinity near $0$ and such that
\begin{equation}\label{eq:LF2}
    \mathbb E\left[F^*\left(\psi_\omega^{(1)}(v,w)_2\right)\right] \leq \tilde \gamma F^*(w),
\end{equation}
for some $\tilde \gamma \in (0,1)$ and every $(v,w)$ in some neighborhood of $(0,0)$. This means that the value of $F^*(w)$ must decrease on average along trajectories. Since $F^*$ is large near the origin, points get ``pushed away" from the set $\{w = 0\}$, which will allow us to prove a lack of convergence in the subsequent section. 

\revision{
To construct $F^*$ we employ a strategy that is inspired by recent advances in fluid dynamics (\citealp{BedrossianBlumenthalPunshon2022,bedrossian2025, BlumenthalCotiGvalani2023}). We construct the function $F^*$ to be a Lyapunov function for the linearized dynamics, that is,
\begin{equation}\label{ineq:LF}
    \mathbb E\left[F^*(\Psi_\omega^{(1)}w)\right] = \gamma F^*(w),~\forall\,w \in \mathbb R^N\setminus\{0\},
\end{equation}
for some $\gamma \in (0,1).$ From this we can conclude (Corollary \ref{cor:LyapFun} below) that \eqref{eq:LF2} holds locally around the origin for any $\tilde \gamma \in (\gamma,1)$. Making the ansatz  
$$F^*(w) = \|w\|^{-p} f^*\left(\frac{w}{\|w\|}\right),$$
with $p>0$ and $f^*: S^{N-1} \to [0,\infty)$, equation \eqref{ineq:LF} is satisfied if and only if 
$$\mathbb E\left[\left\|\Psi^{(1)}_\omega s\right\|^{-p} f^*\left(\frac{\Psi^{(1)}_\omega s}{\left\|\Psi^{(1)}_\omega s\right\|}\right)\right] = \gamma f^*(s),~\forall\,s\in S^{N-1}.$$
}
We will study the family of linear operators $(\mathcal P_q: \mathcal C^0(S^{N-1}) \to \mathcal C^0(S^{N-1}))_{q \in \mathbb R}$ given by
\begin{equation}\label{eq:eigenvalue}
    [\mathcal P_q f] (s) = \mathbb E\left[\left\|\Psi^{(1)}_\omega s\right\|^q f\left(\frac{\Psi^{(1)}_\omega s}{\left\|\Psi^{(1)}_\omega s\right\|}\right)\right].
\end{equation}
Here, $\mathcal C^0(S^{N-1})$ denotes the Banach space of real-valued continuous functions on $S^{N-1}$, equipped with the supremum norm. It can be readily seen that the operators $\mathcal P_q$ are bounded in this norm\footnote{In fact $\|\mathcal P_q\|\leq K_2^q$ for $q \geq 0$ and $\|\mathcal P_q\|\leq K_1^{-q}$ for $q \leq 0$, with $K_1$ and $K_2$ defined by \eqref{eq:K1def} and \eqref{eq:K2def}.}. \revision{Furthermore, the operators $\mathcal P_q$ are positive, that is, they map non-negative functions to non-negative functions. We can reformulate \eqref{eq:eigenvalue} as the eigenvalue problem 
\begin{equation}\label{eq:EV2}
    \mathcal P_{-p} f^* = \gamma f^*,\quad f^*\geq 0,~0<\gamma <1,~p>0.
\end{equation}
Let us pretend for now that the operators $\mathcal P_q$ are compact and strongly positive\footnote{Neither of these properties actually hold.}, that is, they map any non-negative function that is positive somewhere to a function that is positive everywhere. By the Krein-Rutman Theorem \citep{KreinRutman}, for each $q \in \mathbb R$, there would be a unique non-negative eigenfunction $f_q\geq 0$ whose corresponding eigenvalue, say $r(q)$ is simple, isolated, and principal. In this setting, the principal eigenvalue $r(q)$ is closely related to the $q$-th moment Lyapunov exponent $\Lambda_q$ (cf.~Definition \ref{def:MLE}), by the identity $\Lambda_q = \log r(q)$ (see, e.g.~\citealp[Lemma 2]{ArnoldKliemann87}). To find a solution to \eqref{eq:EV2}, we therefore look for a $p>0$ with $\Lambda_{-p} < 0$. In settings where the Krein-Rutman Theorem holds for the operators $\mathcal P_q$, the moment Lyapunov exponent $\Lambda_q$ is known to satisfy $\Lambda_0 = 0$ and $\frac{\dd}{\dd q}\Lambda_q|_{q=0} = \lambda$ \citep[Theorem 1]{ArnoldKliemann87}. If $\lambda>0$, we can choose $0<p\ll 1$ to have $\Lambda_{-p}<0$ and thus a solution to \eqref{eq:EV2}.

If the operators $\mathcal P_q$ are not compact, a similar argument is still possible if one assumes that the operator $\mathcal P_0$ admits a spectral gap. By definition, the operator $\mathcal P_0$ is the Markov operator of the linearized dynamics projected onto the unit sphere. In particular, $\|\mathcal P_0\| = 1$ and the constant function $\mathbf 1 \in C^0(S^{N-1})$ is an eigenfunction with $\mathcal P_0 \mathbf 1 = \mathbf 1$. We say that $\mathcal P_0$ admits a spectral gap if $1$ is a simple eigenvalue and the remaining spectrum is contained in some ball with radius less than 1. This is equivalent to the associated Markov process being uniformly geometrically ergodic (see, e.g.~\citealp{hairerNotes}). Under the spectral gap assumption one can use tools from perturbation theory \citep{Kato} to show that the principal eigenvalue $r(q)$ of $\mathcal P_q$ is analytic in $q$ for a neighborhood of 0. This turns out to be sufficient to find a solution to the eigenvalue problem \eqref{eq:EV2}, whenever $\lambda > 0$ (see \citealp{BedrossianBlumenthalPunshon2022,bedrossian2025, BlumenthalCotiGvalani2023}).
} 

Unfortunately, the discrete nature of our setting prohibits \revision{uniform geometric ergodicity and thus a spectral gap for $\mathcal P_0$}, at least in $\mathcal C^0(S^{N-1})$. In order to circumvent this obstacle, we must consider a different Banach space. For $\alpha \in (0,1)$, we let $\mathcal C^\alpha(S^{N-1})$ denote the Banach space of $\alpha$-Hölder continuous functions on the unit sphere $S^{N-1} \subset \mathbb R^N$, i.e.
$$\mathcal C^\alpha(S^{N-1}) = \left\{f: S^{N-1} \to \mathbb R : \exists\, h>0 \text{ s.t. } |f(s_1)-f(s_2)|\leq h\|s_1-s_2\|^\alpha,~\forall\, s_1,s_2 \in S^{N-1}\right\}$$
and 
$$\|f\|_{C^\alpha} = \|f\|_\infty + \sup_{s_1,s_2 \in S^{N-1}, s_1 \neq s_2} \frac{|f(s_1)-f(s_2)|}{\|s_1-s_2\|^\alpha}.$$
As we will only consider the space of Hölder-continuous functions on $S^{N-1}$ here, we will abbreviate $\mathcal C^\alpha = \mathcal C^\alpha(S^{N-1})$. Also, $L(\mathcal C^\alpha)$ denotes the space of bounded linear operators from $\mathcal C^\alpha$ to itself.

\begin{lemma}[{Proposition V.4.1 in Part A of \citealp{RMP}}]\label{lemm:specGap}
If $\mathcal S(x^*)$ is contracting and strongly irreducible, then there exists an $\alpha \in (0,1)$ such that
\begin{enumerate}
    \item[(i)] there exists an $\hat q> 0$, such that for $q\in (-\hat q, \hat q)$ the operator $\mathcal P_p \in L(\mathcal C^0)$ restricts to a well-defined, bounded operator $\mathcal P_q \in L(\mathcal C^\alpha)$ and the map $\mathcal P_\bullet: (-\hat q,\hat q) \to L(\mathcal C^\alpha)$, $q \mapsto \mathcal P_q $ is analytic
    \item[(ii)] and the operator $\mathcal P_0$ satisfies  
    $$\limsup_{n \to \infty} \|\mathcal P_0^n f -\kappa(f) \mathbf 1\|_{\mathcal C^\alpha}^{\frac{1}{n}}< 1, \forall f \in \mathcal C^\alpha,$$
    for some probability measure $\kappa$ on $S^{N-1}$. Here, we let $\mathbf 1 \in \mathcal C^\alpha$ denote the constant function with value 1.
\end{enumerate}
\end{lemma}
For the proof, we refer to \cite{RMP}. The integrability assumption is satisfied trivially in our setting, as $\Psi^{(1)}_\omega$ can only take finitely many values.

Lemma \ref{lemm:specGap} (ii) says that 1 is an dominant eigenvalue of $\mathcal P_0$, i.e.~it is isolated and the spectral value with the largest absolute value. By classical perturbation theory \citep{Kato} this implies that for $q$ sufficiently close to 0 the largest spectral value of $\mathcal P_q$ is also an isolated eigenvalue and both the dominant eigenvalue and the corresponding Riesz projections are analytic in $q$. 
Put more precisely, we get the following corollary (cf.~also Theorem V.4.3 in Part A of \citealp{RMP}). Here $\dot{\mathcal C}^\alpha$ denotes the dual space of $\mathcal C^\alpha$ and $\mathcal Q^*\in L(\dot{\mathcal C}^\alpha)$ the dual operator of an operator $\mathcal Q \in L(\mathcal C^\alpha)$. 
\begin{corollary}\label{coro:pert}
    In the setting of the previous lemma, there exists a $0 < \tilde q < \hat q$ and analytic maps $r: (-\tilde q, \tilde q) \to \mathbb R$, $f_\bullet: (-\tilde q, \tilde q) \to \mathcal C^\alpha$, $\kappa_\bullet: (-\tilde q, \tilde q) \to \dot {\mathcal C}^\alpha$ and $\mathcal Q_\bullet: (-\tilde q, \tilde q) \to L(\mathcal C^\alpha)$, such that
    \begin{equation}\label{eq:qdecomp}
        \mathcal P_q f = \mathcal Q_q f  + r(q)\langle\kappa_q,f\rangle f_q,
    \end{equation}
    where $\mathcal Q_q f_q = 0$, $\mathcal Q_q^* \kappa_q = 0$, $\langle \kappa_q, f_q\rangle = 1$ and 
    $$\limsup_{n \to \infty} \left\|\mathcal Q_q^n\right\|^{\frac{1}{n}} < r(p).$$
    Furthermore, $r(0) = 1$, $f_0 = \mathbf 1$ and $\kappa_0 \in \dot{\mathcal C}^\alpha$ is given by $\langle\kappa_0, f\rangle = \kappa(f)$, where $\kappa$ is the probability measure from Lemma \ref{lemm:specGap} (ii).
\end{corollary}
The arguments for the rest of the section are similar to the ones made in chapter 4 of \cite{BedrossianBlumenthalPunshon2022}. Henceforth, let $\alpha\in (0,1)$ be as in Lemma \ref{lemm:specGap}.
\begin{theorem}\label{theo:LyapFun}
    Let $x^* \in \mathcal M$ be regular with $\lambda(x^*) > 0$. There exist constants $p>0$, $\gamma \in (0,1)$ and $\alpha \in (0,1)$ and a positive function $f^*\in \mathcal C^\alpha$ such that for all $s \in S^{N-1}$ we have
        \begin{equation}\label{eq:efTwist}
            \mathbb E\left[\left\|\Psi_\omega^{(1)} s\right\|^{-p}f^*\left(\frac{\Psi_\omega^{(1)} s}{\left\|\Psi_\omega^{(1)} s\right\|}\right)\right] = \gamma f^*(s).
        \end{equation}
\end{theorem}
\begin{proof}
Note that \eqref{eq:efTwist} states that $\mathcal P_{-p}f^* = \gamma f^*$.
In the following, we will show that $\frac{\dd}{\dd q} r(q)|_{q = 0} = \lambda(x^*)>0$. This will imply that for sufficiently small $p>0$, we can set $\gamma := r(-p)<1$ and $f^*:= f_{-p}$. Since, for small $p$, the function $f_{-p}$ is close\footnote{in the $\mathcal C^\alpha$-sense, but thus in particular in the $\mathcal C^0$-sense} to $f_0 = \mathbf 1$, it is indeed a positive function and by \eqref{eq:qdecomp}, we have $\mathcal P_{-p}f^* = \gamma f^*$. 

In order to show $\frac{\dd}{\dd q} r(q)|_{q = 0} = \lambda(x^*)$, note that for $q$ sufficiently close to 0, we have $\langle \kappa_q, \mathbf 1\rangle = \langle \kappa_q, f_0\rangle \neq 0$. By Corollary \ref{coro:pert}, this allows us to express $r(q)$ by
$$\log r(q) = \lim_{n \to \infty} \frac{1}{n} \log\|\mathcal P_q^n \mathbf 1\|_{\mathcal C^\alpha}.$$
Using Jensen's inequality, we can estimate
\begin{align*}
    \log r(q) &= \lim_{n \to \infty} \frac{1}{n} \log\|\mathcal P_q^n \mathbf 1\|_{\mathcal C^\alpha} \geq \lim_{n \to \infty} \frac{1}{n} \log \|\mathcal P_q \mathbf 1\|_{\mathcal C^0}\\
    &=\lim_{n \to \infty} \frac{1}{n} \log\left(\sup_{s \in S^{N-1}}\mathbb E \left[\left\|\Psi^{(1)}_\omega s\right\|^{q}\right]\right)\\
    &=\lim_{n \to \infty} \frac{1}{n} \log\left(\sup_{s \in S^{N-1}}\mathbb E \left[e^{q\log\left\|\Psi^{(1)}_\omega s\right\|}\right]\right)\\
    &\geq q\lim_{n \to \infty} \frac{1}{n} \sup_{s \in S^{N-1}}\mathbb E \left[\log\left\|\Psi^{(1)}_\omega s\right\|\right].
\end{align*}
By the Oseledec theorem (cf.~Theorem \ref{theo:Oseledets}) the limit in the last line is equal to $\lambda(x^*)$. Thus we get $r(q) \geq e^{q\lambda(x^*)}$, for sufficiently small $q$. Since we already know that $r$ is differentiable in 0 from Corollary \ref{coro:pert} and since $r(0) = 1$, this implies $\frac{\dd}{\dd q} r(q)|_{q = 0} = \lambda(x^*)$, completing the proof.
\end{proof}
We define a function $F^*:  \mathbb R^N\setminus \{0\} \to \mathbb R_{>0}$ by 
\begin{equation}\label{eq:FstarDef}
    F^*(w) := \|w\|^{-p} f^*\left(\frac{w}{\|w\|}\right).
\end{equation}

\begin{corollary}\label{cor:linLyapFun}
We have
\begin{equation}\label{eq:linLyapFun}
    \mathbb E\left[F^*(\Psi_\omega^{(1)}w)\right] = \gamma F^*(w).
\end{equation}
\end{corollary}
\begin{proof}
This is a direct consequence of \eqref{eq:efTwist}. Let $w \in \mathbb R^N \setminus \{0\}$ and set $s := \frac{w}{\|w\|} \in S^{N-1}$. Then
\begin{align*}
    \mathbb E\left[F^*(\Psi_\omega^{(1)}w)\right] &=  \mathbb E\left[\left\|\Psi_\omega^{(1)} w\right\|^{-p}f^*\left(\frac{\Psi_\omega^{(1)} w}{\|\Psi_\omega^{(1)} w\|}\right)\right]\\
    &=  \|w\|^{-p}\mathbb E\left[\left\|\Psi_\omega^{(1)} s\right\|^{-p}f^*\left(\frac{\Psi_\omega^{(1)} s}{\|\Psi_\omega^{(1)} s\|}\right)\right]\\
    &= \|w\|^{-p}\gamma f^*(s) = \gamma F^*(w),
\end{align*}
showing the claim.
\end{proof}
This establishes that $F^*$ is a Lyapunov function for the linearized process induced by $\Psi$. For the remainder of this section, we will show that, in a neighborhood of the origin, $F^*$ is also a Lyapunov function for the Markov process induced by the nonlinear cocycle $\psi$.
Since $f^*$ is continuous, positive, and has a compact domain, it is both bounded and bounded away from zero, i.e.~we can find constants $0<C_-\leq C_+<\infty$ such that
\begin{equation}\label{ineq:fpBound}
    C_- \leq f^*(s) \leq C_+ ,~\forall s \in S^{N-1}.
\end{equation}
As a direct consequence, we also get the bound
\begin{equation}\label{ineq:VpBound}
    C_- \|w\|^{-p}\leq F^*(w) \leq C_+\|w\|^{-p} ,~\forall w \in \mathbb R^N \setminus \{0\}.
\end{equation}

\begin{lemma}
    For every $\varepsilon>0$, there exists a $\delta>0$ such that for $w, \tilde w \in \mathbb R^N\setminus \{0\}$ we have
    \begin{equation}\label{impl:VpCont}
        \frac{\|w-\tilde w\|}{\|w\|}< \delta ~\Rightarrow~ \frac{\left|F^*(w)-F^*(\tilde w)\right|}{F^*(w)}< \varepsilon. \end{equation}
\end{lemma}
\begin{proof}
    Using $\alpha$-Hölder continuity of $f^*$ and the bounds in \eqref{ineq:fpBound}, we can estimate
    \begin{align*}
        \left|F^*(w)-F^*(\tilde w)\right| &= \left|\|w\|^{-p} f^*\left(\frac{w}{\|w\|}\right)-\|\tilde w\|^{-p} f^*\left(\frac{\tilde w}{\|\tilde w\|}\right)\right|\\
        &\leq \|w\|^{-p}\left| f^*\left(\frac{w}{\|w\|}\right)- f^*\left(\frac{\tilde w}{\|\tilde w\|}\right)\right| + \big|\|w\|^{-p}-\|\tilde w\|^{-p}\big| f^*\left(\frac{\tilde w}{\|\tilde w\|}\right)\\
        &\leq F^*(w) C_-^{-1}\|f^*\|_{\mathcal C^\alpha}\left\| \frac{w}{\|w\|}- \frac{\tilde w}{\|\tilde w\|}\right\|^\alpha + F^*(w) C_+ C_-^{-1} \left|1-\left(\frac{\|\tilde w\|}{\|w\|}\right)^{-p}\right|.
    \end{align*}
    In order to complete the proof, we will show that for each $\varepsilon>0$ there exists a $\delta >0$, such that for $w, \tilde w \in \mathbb R^N\setminus \{0\}$ with $\|w\|^{-1}\|w-\tilde w\| < \delta $ we have
    \begin{align}
        \label{ineq:linnonlin1}
        \left\| \frac{w}{\|w\|}- \frac{\tilde w}{\|\tilde w\|}\right\| &\leq \left(\frac{C_- \varepsilon}{\|f^*\|_{\mathcal C^{\alpha}}}\right)^{\frac{1}{\alpha}} =: \varepsilon_1&&\text{ and } & \left|1-\left(\frac{\|\tilde w\|}{\|w\|}\right)^{-p}\right| &\leq \frac{C_+ \varepsilon}{C_-} =: \varepsilon_2.
    \end{align} 
   We can bound
   \begin{align*}
       \left\| \frac{w}{\|w\|}- \frac{\tilde w}{\|\tilde w\|}\right\| &=\|w\|^{-1}\left\|w-\tilde w + \left(1-\frac{\|w\|}{\|\tilde w\|}\right)\tilde w\right\|\\
       &\leq \frac{\|w-\tilde w\|}{\|w\|} + \frac{\big|\|w\|-\|\tilde w\|\big|}{\|w\|} \leq 2 \frac{\|w-\tilde w\|}{\|w\|}
   \end{align*}
    to see that the first inequality in \eqref{ineq:linnonlin1} is satisfied for $w, \tilde w$ with ${\|w\|^{-1}\|w-\tilde w\| < \delta_1 := \frac{1}{2}\varepsilon_1}$. Since the map $t \mapsto t^{-p}$ is continuous at $t = 1 \mapsto 1$, there also exists a $\delta_2>0$ such that
    $$\left|1-\frac{\|\tilde w\|}{\|w\|}\right|\leq\frac{\|w-\tilde w\|}{\|w\|} < \delta_2 ~\Rightarrow~ \left|1-\left(\frac{\|\tilde w\|}{\|w\|}\right)^{-p}\right| < \varepsilon_2.$$
    Thus \eqref{impl:VpCont} will be satisfied for $\delta = \min(\delta_1, \delta_2).$
\end{proof}

\begin{lemma}\label{lemm:linnonlin}
For every $\varepsilon > 0$, there exists a $\delta >0$ such that
\begin{equation}\label{ineq:linnonlinlemm}
    F^*\left(\psi_\omega^{(1)}(v,w)_2\right) \leq (1+\varepsilon) F^*\left(\Psi_\omega^{(1)}w\right),
\end{equation}
for all $\omega \in \Omega$ and $(v,w) \in V_\delta$, where $V_\delta$ is the neighborhood given in Lemma \ref{lemm:approx}/Corollary \ref{coro:approx}.
\end{lemma}
\begin{proof}
Let $\varepsilon >0$ and choose $\tilde\delta>0$ such that the conclusion \eqref{impl:VpCont} of the previous lemma holds. Recall that (cf.~\eqref{eq:K1def})
\begin{equation*}\label{eq:defK}
    K_1 := \inf_{\omega \in \Omega} \left\|\left(\Psi^{(1)}_\omega\right)^{-1}\right\|^{-1} >0
\end{equation*}
and let $\delta = K_1\tilde \delta$. Then, by Lemma \ref{lemm:approx}, we have
$$\left\|\Psi^{(1)}_\omega w-\psi^{(1)}_\omega(v,w)\right\| \leq \delta \|w\| \leq K_1^{-1} \delta \left\|\Psi^{(1)}_\omega w\right\| = \tilde \delta \left\|\Psi^{(1)}_\omega w\right\|,$$
for all $\omega \in \Omega$ and $(v,w) \in V_\delta$.
By \eqref{impl:VpCont}, this implies
\begin{align*}
    \frac{\left|F^*\left(\Psi^{(1)}_\omega w\right)-F^*\left(\psi^{(1)}_\omega(v,w)\right)\right|}{F^*\left(\Psi^{(1)}_\omega w\right)} < \varepsilon,
\end{align*}
which implies \eqref{ineq:linnonlinlemm}.
\end{proof}

\begin{corollary}\label{cor:LyapFun}
For every $\varepsilon>0$, there exists a $\delta >0$ s.t. for all $\omega \in \Omega$ and all ${(v,w) \in V_\delta \setminus (\mathbb R^{D-N}\times \{0\})}$ we have $\psi_\omega^{(1)}(v,w)_2 \neq 0$ and
$$\mathbb E\left[F^*\left(\psi_\omega^{(1)}(v,w)_2\right)\right] \leq \left(\gamma+\varepsilon\right)F^*(w).$$
\end{corollary}
In particular, if we choose $0<\varepsilon<1-\gamma$ this shows that $F^*$ is indeed a Lyapunov function on a neighborhood of the origin. 
 \begin{proof}
 First note that if we choose $\delta<K_1$, using Corollary \ref{coro:approx}, we get 
 $$\left\|\psi_\omega^{(1)}(v,w)_2\right\|\geq \left\|\Psi_\omega^{(1)}w\right\| - \left\|\Psi_\omega^{(1)}w - \psi_\omega^{(1)}(v,w)_2\right\|\geq K_1\|w\| - \delta \|w\| > 0$$
 for all $\omega \in \Omega$ and all ${(v,w) \in V_\delta\setminus(\mathbb R^{D-N}\times \{0\})}$. Let $\varepsilon>0$ and choose $\delta>0$ small enough, such that both $\delta< K_1$ and that the conclusion of Lemma \ref{lemm:linnonlin} holds. Together with Corollary \ref{cor:linLyapFun} this yields
 \begin{align*}
     \mathbb E\left[F^*(\psi_\omega^{(1)}(v,w)_2)\right] \leq (1+\varepsilon) \mathbb E\left[F^*(\Psi_\omega^{(1)}w)\right]
     =(1+\varepsilon)\gamma F^*(w) \leq (\gamma+\varepsilon) F^*(w),
 \end{align*}
 for all $\omega \in \Omega$ and all ${(v,w) \in V_\delta\setminus(\mathbb R^{D-N}\times \{0\})}$.
 \end{proof}

\subsection{Stochastic Gradient Descent - the Unstable Case}\label{sec:sgdunstable}
\begingroup
\def\thetheorem{\ref{theo:mainSGD} (ii)}
\begin{theorem}
Let $x^* \in \mathcal M$ be regular with $\lambda(x^*) > 0$. Then $x^* \notin \operatorname{supp}(X_{\lim}^{\operatorname{SGD}})$. 
\end{theorem}
\addtocounter{theorem}{-1}
\endgroup

\begin{proof}[of Theorem \ref{theo:mainSGD} (ii)]
    Suppose $x^* \in \mathcal M$ is regular with $\lambda(x^*) > 0$. In the following, we will show that there exists a $\delta > 0$ such that the neighborhood $0 \in V_\delta\subseteq \mathbb R^D$ from Lemma \ref{lemm:approx}/Corollary \ref{coro:approx} satisfies
    \begin{equation}\label{eq:leaveVdelt}
        \mathbb P\left(\exists n\, \in \mathbb N, \text{ s.t. } \psi_\omega^{(n)}(v,w) \notin V_\delta  \right) = 1, ~\forall, (v,w) \in V_\delta \setminus \left(\mathbb R^{D-N} \times \{0\}\right).
    \end{equation}
    We first argue why this is sufficient to show $x^* \notin \operatorname{supp}(X_{\lim}^{\operatorname{SGD}}).$ Let $x^* \in U \subseteq \mathcal M$ be the neighborhood given by
    $$U := \chi(V_\delta) \cap \mathcal M = \chi\left(V_\delta \cap \left(\mathbb R^{D-N} \times \{0\}\right) \right).$$
    Suppose for now that $\omega$ is such that $X_{\lim}^{\operatorname{SGD}} \in U$. By the openness of $\chi(V_\delta)$, there then either exists an $n \in \mathbb N_0$ s.t.~$X_n^{\operatorname{SGD}} \in U$ or there exists an $m \in \mathbb N_0$, such that $X_{m+n}^{\operatorname{SGD}} \in \chi(V_\delta) \setminus \mathcal M$ for all $n \in \mathbb N_0$. 
    By Lemma \ref{lemm:nullsets}, the former happens with probability zero. The probability for the latter to happen can be estimated by
    \begin{align*}
        &\phantom{=}~\mathbb P\left(\exists\, m \in \mathbb N_0, ~ \forall\, n \in \mathbb N_0, ~ X_{m+n}^{\operatorname{SGD}}(\omega) \in \chi(V_\delta) \setminus \mathcal M \right)\\
        &\leq \sum_{m = 1}^\infty \mathbb P\left(\forall\, n \in \mathbb N_0, ~ \varphi_{\theta^m \omega}^{(n)}(X_m^{\operatorname{SGD}}(\omega)) \in \chi(V_\delta) \setminus \mathcal M\right)\\  
        &= \sum_{m = 1}^\infty \mathbb P\left(\forall\, n \in \mathbb N_0, ~ \psi_{\theta^m \omega}^{(n)}(\chi^{-1}( X_m^{\operatorname{SGD}}(\omega))) \in V_\delta \setminus (\mathbb R^{D-N} \times \{0\})\right).
    \end{align*}
    Since $X_m^{\operatorname{SGD}}$ only depends on $\omega_{\operatorname{init}}$ and $\omega_1, \dots, \omega_m$ and the random map $\psi_{\theta^m \omega}^n$ only depends on $\omega_{m+1}, \dots \omega_{m+n}$, they are independent. Thus, if \eqref{eq:leaveVdelt} holds, this probability will also be zero.

    It remains to find a $\delta > 0$ such that \eqref{eq:leaveVdelt} holds. Let $p > 0$, $\gamma \in (0,1)$ and $f^*: S^{N-1} \to (0, \infty)$ be such that the conclusion of Theorem \ref{theo:LyapFun} holds. Also, let $F^*$ be the function defined in \eqref{eq:FstarDef}, choose some $0<\varepsilon < 1-\gamma$ and let $\delta$ be such that the conclusion of Corollary \ref{cor:LyapFun} holds. Without loss of generality, we assume $V_\delta \subseteq \mathbb R^{D-N} \times \mathcal B_{\mathbb R^N}(1)$\footnote{If not, consider the intersection of $V_\delta$ and $\mathbb R^{D-N} \times \mathcal B_{\mathbb R^N}(1)$ henceforth.}. Recall that $\tau_{\delta,\omega}: V_\delta \to \mathbb N \cup \{\infty\}$ is given by 
    $$\tau_{\delta, \omega}(v,w) := \inf\left\{n \in \mathbb N: \psi^{(n)}_\omega(v,w) \notin V_\delta\right\}.$$
    We can reformulate \eqref{eq:leaveVdelt} as 
    $$\mathbb P\left(\tau_{\delta, \omega}(v,w) < \infty \right) = 1, ~\forall (v,w) \in V_\delta \setminus \left(\mathbb R^{D-N} \times \{0\}\right).$$
    Using Corollary \ref{cor:LyapFun} inductively, one can show that
    $$\mathbb E \left[ \mathds 1_{\tau_{\delta, \omega}(v,w) \geq n} F^*\left(\psi_\omega^{(n)}(v,w)_2\right)\right] \leq (\gamma+\varepsilon)^nF^*(w),$$
    for all $n \in \mathbb N$ and $(v,w) \in V_\delta$. Since $V_\delta \subseteq \mathbb R^{D-N} \times \mathcal B_{\mathbb R^N}(1)$, we have
    $$F^*(w) \geq C_-\|w\|^{-p} \geq C_-, ~\forall (v,w) \in V_\delta.$$
    This allows us to compute
    \begin{align*}
        \mathbb P(\tau_{\delta, \omega}(v,w) = \infty) &= \lim_{n \to \infty} \mathbb P(\tau_{\delta, \omega}(v,w) > n)\\
        &\leq C_-^{-1}\lim_{n \to \infty} \mathbb E\left[ \mathds 1_{\tau_{\delta, \omega}(v,w) > n} F^*\left(\psi_\omega^{(n)}(v,w)_2\right)\right]\\
        &\leq C_-^{-1} \lim_{n \to \infty} (\gamma + \varepsilon)^nF^*(w) = 0,
    \end{align*}
    for all $(v,w)\in V_\delta \setminus(\mathbb R^{D-N} \times \{0\})$.
\end{proof}

\section*{Acknowledgements}
The authors would like to thank Tim van Erven for his insightful feedback. 
Furthermore, the authors thank the DFG SPP 2298 for supporting their research. Both authors have been additionally supported by Germany’s Excellence Strategy – The Berlin Mathematics Research Center MATH+ (EXC-2046/1, project ID:
390685689), in the case of D.C. via the Berlin Mathematical School and in the case of M.E. via projects AA1-8, AA1-18 and EF45-5. Furthermore, M.E. thanks the DFG CRC 1114, the Einstein Foundation and the Dutch Research Council NWO (VI.Vidi.233.133) for support.
\bibliography{biblio}
\appendix

\section{Comparison to second moment linear stability}\label{app:WuMaW}
A different notion of linear stability for SGD was introduced by \cite{WuMaE}. Adapted to the notation introduced in Section \ref{sec:RDS}, their condition can be expressed as follows.
\begin{definition}[Definition 2 in \citealp{WuMaE}]\label{def:2ndmom}
    A global minimum $x^* \in \mathcal M$ is called \emph{second moment linearly stable}, if there exists a constant $C$ such that 
    $$\mathbb E\left[\left\|\Phi^{(n)}_\omega x \right\|^2\right] \leq C \|x\|^2,$$
    for all $x \in \mathbb R^D$ and all $n \in \mathbb N$. Here $\Phi^{(n)}_\omega$ denotes the linearization around $x^*$ of $n$ steps of SGD with seed $\omega$ as defined in \eqref{def:Phi}.
\end{definition}
In contrast, we call a global minimum $x^*$ with $\lambda(x^*) < 0$ \emph{almost surely stable}. As will be argued below, second moment linear stability is almost a strictly stronger condition then almost sure stability. The fact that second moment stability and almost-sure stability are in general not equivalent, can already be seen for 1-dimensional linear stochastic processes. Let $(X_n)_{n \in \mathbb N_0}$ be the stochastic process given by $X_0 = 1$ and
$$X_{n+1} = Y_{n+1}X_n,$$
where $(Y_n)_{n \in \mathbb N}$ is an i.i.d. sequence of real-valued random variables with ${\mathbb E[\log_+ |Y_1|] < \infty}$.\footnote{Here $\log_+ |Y_1| = \max (0, \log |Y_1|)$.} By the strong law of large numbers, we have 
$$\lim_{n \to \infty} \frac{1}{n} \log |X_n| = \lim_{n \to \infty} \frac{1}{n} \sum_{k = 1}^n\log |Y_n| = \mathbb E[\log |Y_1|] \in [-\infty, \infty),~\text{almost surely}.$$
Thus, if $\mathbb E[\log |Y_1|] < 0$, then the linear process $(X_n)$ is almost-surely stable, i.e.~$X_n \to 0$ with probability 1. At the same time, since $X_n$ and $Y_{n+1}$ are independent we have
$$\mathbb E\left[|X_{n+1}|^2 \right] = \mathbb E\left[|Y_{n+1}|^2 \right]\mathbb E\left[|X_{n}|^2 \right] = \mathbb E\left[|Y_{1}|^2 \right]\mathbb E\left[|X_{n}|^2 \right]$$
and therefore $\mathbb E\left[|X_{n}|^2 \right] = \mathbb E\left[|Y_{1}|^2 \right]^n$. Thus $(X_n)$ is second moment stable, i.e.~there exists a $C>0$ such that $\mathbb E\left[|X_{n}|^2 \right] < C$, if and only if $\mathbb E\left[|Y_{1}|^2 \right]\leq 1$. However, while $\mathbb E\left[|Y_{1}|^2 \right]\leq 1$ implies $\mathbb E[\log |Y_1|] < 0$ by Jensen's inequality, the converse is clearly not true.\footnote{Suppose for example that $Y_1$ takes the values $\frac{1}{3}$ and $2$ with probability $\frac{1}{2}$ each. A simple calculation shows $\mathbb E[\log |Y_1|] = \frac{1}{2}(\log(2)-\log(3)) <0$, while $\mathbb E\left[|Y_{1}|^2 \right] = \frac{19}{9} > 1$.}

Second moment stability is closely related to so-called moment Lyapunov exponents (see e.g.~\citealp{momentLE}), which have been linked to central limit theorems as well as large deviation theory for the convergence of finite-time Lyapunov exponents (see e.g.~\citealp{RMP,ArnoldKliemannOeljeklaus86,ArnoldKliemann87}).
\begin{definition}\label{def:MLE}
    For any $p \in \mathbb R$, the $p$-th moment Lyapunov exponent $\Lambda_p(x^*)$ of a global minimum $x^* \in \mathcal M$ is given by
    $$\Lambda_{p}(x^*) := \sup_{w \in \mathbb R^N}\lim_{n \to \infty} \frac{1}{n} \log  \mathbb E \left[\left\|\Psi_\omega^{(n)}w\right\|^p\right].$$
\end{definition}
Convergence for each $w \in \mathbb R^N$ follows from a subadditivity argument.
The precise relation between second moment Lyapunov exponents and second order stability can be stated as follows.
\begin{proposition}\label{prop:secondM}In order for a global minimum $x^* \in \mathcal M$ to be second moment linearly stable, it is
    \begin{enumerate}  
        \item[(i)] necessary to have $\Lambda_{2} \leq 0$  and 
        \item[(ii)] sufficient to have $\Lambda_{2} < 0$.
    \end{enumerate}
\end{proposition}
\begin{proof}
    Note that by \eqref{eq:PhiViaPsi}, a global minimum $x^*$ is second moment linearly stable if and only if there exists a $C>0$ such that
    $$\mathbb E\left[\left\|\Psi^{(n)}_\omega w \right\|^2\right] \leq C \|w\|^2,$$
    for all $w \in \mathbb R^N$ and all $n \in \mathbb N$. Now (i) follows directly from the definition of the second moment Lyapunov exponent. In order to prove (ii), suppose $\Lambda_{2,w} < 0$ for all $w \in \mathbb R^N$ and thus in particular for the unit vectors $e_1, \dots, e_N$. Thus 
    $$\lim_{n \to \infty} \mathbb E \left[\left\|\Psi_\omega^{(n)}e_i\right\|^2\right] = 0, ~\forall\, i \in [N].$$
    Let $C>0$ be given by
    $$C = \sup_{i \in [N], \, n \in \mathbb N} \mathbb E \left[\left\|\Psi_\omega^{(n)}e_i\right\|^2\right] < \infty.$$
    Now for $w = w_1 e_1 + \dots + w_N e_N$, we have
    \begin{align*}
        \mathbb E\left[\left\|\Psi^{(n)}_\omega w \right\|^2\right] &= \mathbb E\left[\left\|\Psi^{(n)}_\omega \left(\sum_{i = 1}^N w_ie_i\right)\right\|^2\right] = \mathbb E\left[\left\|\sum_{i = 1}^N w_i\Psi^{(n)}_\omega e_i\right\|^2\right]\\
        &\leq \mathbb E\left[\left(\sum_{i = 1}^N w_i\left\|\Psi^{(n)}_\omega e_i\right\|\right)^2\right]\leq \mathbb E\left[N\sum_{i = 1}^N w_i^2\left\|\Psi^{(n)}_\omega e_i\right\|^2\right]\\
        &\leq NC \sum_{i = 1}^N w_i^2 = NC\|w\|^2.
    \end{align*}
\end{proof}
\begin{proposition}
    For each $p \in \mathbb R$, we have $\Lambda_p(x^*) \geq p \lambda(x^*)$. In particular $\Lambda_2(x^*)\leq 0$ implies $\lambda(x^*) \leq 0$.
\end{proposition}
\begin{proof}
    As a consequence of Oseledec theorem (cf.~ Theorem \ref{theo:Oseledets}), we have
    $$\lambda(x^*) = \sup_{w \in \mathbb R^N} \lim_{n \to \infty} \frac{1}{n} \mathbb E\left[\log\left\|\Psi^{(n)}_\omega w \right\|\right].$$
    The proposition follows from Jensen's inequality.
\end{proof}
To summarize, if $x^* \in \mathcal M$ is second moment stable in the sense of \cite{WuMaE}, it satisfies $\Lambda_2(x^*) \leq 0$ and thus $\lambda(x^*) \leq 0$. Most of these points should be regular in the sense of Definition \ref{def:regular} and even satisfy $\lambda(x^*) < 0$. By Theorem \ref{theo:mainSGD}, these global minima lie in the support of $X_{\lim}^{\operatorname{SGD}}$. On the other hand the one dimensional example demonstrated above suggests that it is possible to have global minima $x^* \in \mathcal M$ with $\lambda(x^*) < 0$ and $\Lambda_2(x^*) > 0$. Given that they are regular, these global minima will be in the support of $X_{\lim}^{\operatorname{SGD}}$, but will not be second moment linearly stable in the sense of \cite{WuMaE}. Notably the paper by \cite{WuMaE} contains a sufficient condition for second order linear stability (cf.~ Therorem 1 in \citealp{WuMaE}). By the argument above, their condition is also a sufficient condition for $\lambda(x^*) \leq 0$. 

In \cite{maYing}, a more general notion of ``$k$-th order linear stability" is introduced. If $p = k$ is an even natural number this is equivalent to the following generalization of second moment linear stability (cf.~Remark 1 in \citealp{maYing}). 
\begin{definition}
    For $p>0$, a global minimum $x^* \in \mathcal M$ is called $p$-th moment linearly stable, if there exists a constant $C$ such that 
    $$\mathbb E\left[\left\|\Phi^{(n)}_\omega x\right\|^p\right] \leq C\|x\|^p,$$
    for all $x \in \mathbb R^D$ and all $n \in \mathbb N$.
\end{definition}
Proposition \ref{prop:secondM} can be extended to $p$-th moment linear stability mutatis mutandis.

\end{document}